\documentclass[sigconf,table]{acmart}

\copyrightyear{2022}
\acmYear{2022}
\setcopyright{rightsretained}

\acmConference[SIGIR '22]{Proceedings of the 45th International ACM SIGIR Conference on Research and Development in Information Retrieval}{July 11--15, 2022}{Madrid, Spain.}
\acmBooktitle{Proceedings of the 45th Int'l ACM SIGIR Conference on Research and Development in Information Retrieval (SIGIR '22), July 11--15, 2022, Madrid, Spain}
\acmDOI{10.1145/3477495.3531722}
\acmISBN{978-1-4503-8732-3/22/07}

\usepackage{etoolbox}
\makeatletter
\patchcmd{\maketitle}{\@copyrightpermission}{
   \begin{minipage}{0.3\columnwidth}
     \href{http://creativecommons.org/licenses/by/4.0/}{\includegraphics[width=0.90\textwidth]{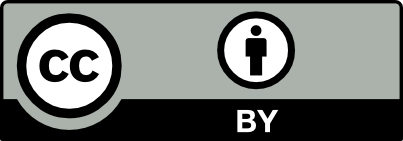}}
   \end{minipage}\hfill
   \begin{minipage}{0.7\columnwidth}
     \href{http://creativecommons.org/licenses/by/4.0/}{This work is licensed under a Creative Commons Attribution International 4.0 License.}
   \end{minipage}

   \vspace{5pt}
}{}{}

\makeatother

\usepackage{enumitem}
\newlist{inlinelist}{enumerate*}{1}
\setlist*[inlinelist,1]{%
  label=(\roman*),
}
\usepackage{subfigure}

\newcommand{\fdcomment}[1]{\textcolor{red}{[FD: #1]}}

\renewcommand{\fdcomment}[1]{\textcolor{red}{[FD: #1]}}

\title{Retrieval-Enhanced Machine Learning}

\author{Hamed Zamani}
\authornote{Both authors contributed equally to the paper.}
\affiliation{\institution{University of Massachusetts Amherst}}
\email{zamani@cs.umass.edu}

\author{Fernando Diaz}
\authornotemark[1]
\affiliation{\institution{Google Research}}
\email{diazf@acm.org}

\author{Mostafa Dehghani}
\affiliation{\institution{Google Research}}
\email{dehghani@google.com}

\author{Donald Metzler}
\affiliation{\institution{Google Research}}
\email{metzler@google.com}

\author{Michael Bendersky}
\affiliation{\institution{Google Research}}
\email{bemike@google.com}

\begin{document}

\fancyhead{}

\begin{abstract}

    Although information access systems have long supported \textit{people} in accomplishing a wide range of tasks, we propose broadening the scope of users of information access systems to include task-driven \emph{machines}, such as machine learning models.  In this way, the core principles of indexing, representation, retrieval, and ranking can be applied and extended to substantially improve model generalization, scalability, robustness, and interpretability.  We describe a generic retrieval-enhanced machine learning (REML) framework, which includes a number of existing models as special cases. REML challenges information retrieval conventions, presenting opportunities for novel advances in core areas, including optimization.  The REML research agenda lays a foundation for a new style of information access research and paves a path towards advancing machine learning and artificial intelligence.
\end{abstract}



\maketitle

\section{Introduction}
\label{sec:introduction}
The vast majority of existing machine learning (ML) systems are designed to be self-contained, with both knowledge and reasoning encoded in model parameters. Consequently, increasing the capacity of machine learning models by increasing their parameter size generally leads to higher accuracy \cite{Goodfellow-et-al-2016}. For example, the number of parameters used in state-of-the-art language models has increased from 94 million in ELMo~\cite{peters2018elmo} to 1.6 trillion in Switch Transformers~\cite{Fedus2021t5xxl}, an over $16\times$ increase in just three years (2018 -- 2021). 
Despite these successes, improving performance by increasing the number of model parameters can incur significant cost and limit access to a handful of organizations that have the resources to train them \cite{bender:parrots}.  As such, focusing model development on the number of parameters is neither scalable nor sustainable in the long run. 


Motivated by recent work demonstrating both that high capacity models memorize training data \cite{carlini21extracting} and that using retrieval-style methods can offload memorization to storage \cite{borgeaud:retro}, we propose the augmenting ML models with access to stored information through information retrieval (IR) techniques. Whereas IR has proven an effective tool to support people accessing large text corpora, we believe that IR can be extended to support machines accessing not just large text corpora but more abstractly-represented knowledge stores.  By designing machine learning architectures that have explicit access to an information retrieval system, we can decouple reasoning from memory, reducing the required model parameters and leveraging the efficiency, scalability, and effectiveness of IR techniques.  We refer to this class of approaches as \textit{retrieval-enhanced machine learning} (REML).  In this paper, we describe how core principles of indexing, representation, retrieval, and ranking can be used to develop REML models.

Using retrieval to improve model accuracy is not without precedent.  Predating modern machine learning methods, the IR community developed some of the earliest known retrieval-enhanced machine learning models. For example, pseudo-relevance feedback \cite{Attar:1978,Croft:1979} leverages a retrieval system to analyze results of an `initial' search query before producing a final ranking.  This purely algorithmic use of a retrieval system in order to improve ranking model performance foreshadows its usefulness in modern applications.
More recently, natural language processing models that incorporate retrieval capabilities have been shown to improve model performance~\cite{Lewis+al:2020,Guu+al:2020}. Although leveraging rather basic retrieval models, these approaches present an opportunity for ML systems to be further improved with more sophisticated IR methods.

We introduce a generic framework that enables ML models to be augmented with IR capabilities that support querying a corpus for useful information, utilizing  retrieved results, providing feedback to the retrieval model, and, if necessary, storing information for future access. This framework is flexible enough to both represent several existing ML models and scaffold future models.

This paper is organized in order to motivate, describe, and ground REML as a research program.  We begin in Section~\ref{sec:motivation} by describing the motivation for REML, specifically demonstrating why IR techniques provide a unique opportunity for ML.  In Section~\ref{sec:framework}, we discuss the challenges in developing each component of the proposed framework and suggest three categories of optimization approaches for REML models: (1) independent optimization of prediction and retrieval models, (2) their conditional optimization, and (3) their joint end-to-end optimization. Using this framework, in Section~\ref{sec:case_studies}, we review  several existing ML models in order to draw connections to REML. And, although these related models suggest the potential benefit of REML, substantial open research questions limit the applicability and effectiveness of contemporary IR methods.  In Section~\ref{sec:research-agenda}, we conclude with a broad research program in REML, touching on the opportunity for the different subareas of IR research to contribute to the advancement of ML model performance.

\begin{figure*}[t]
    \centering
    \vspace{-0.4cm}
    \subfigure[Cat 1: Retrieval-only]{\label{fig:reml:a}        \includegraphics[trim={4cm 7.5cm 13cm 0cm},clip,width=.24\textwidth]{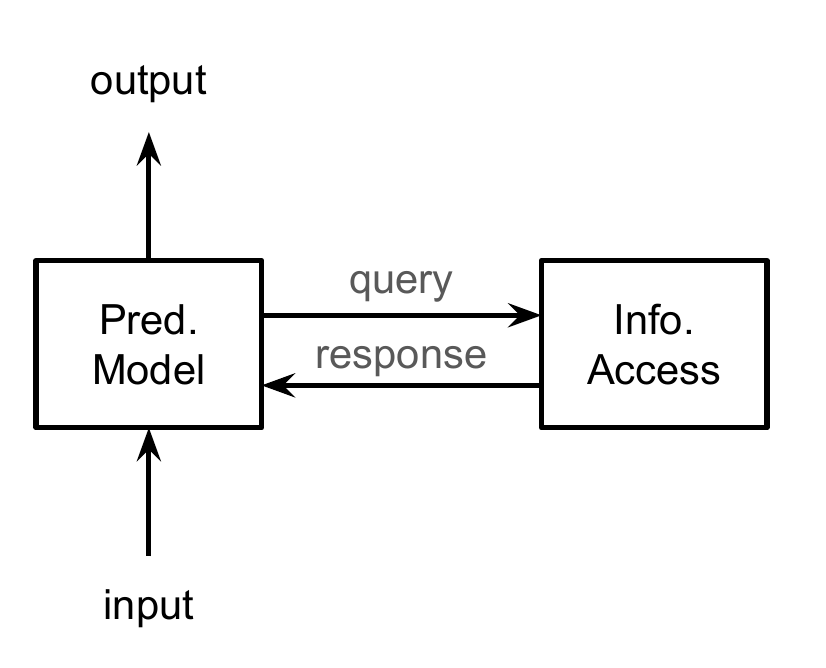}}
    \subfigure[Cat 2: Retrieval with memory]{\label{fig:reml:b}        \includegraphics[trim={13cm 7.5cm 4cm 0cm},clip,width=.24\textwidth]{XX-graphics/REML.pdf}}
    \subfigure[Cat 3: Retrieval with feedback]{\label{fig:reml:c}        \includegraphics[trim={4cm 0cm 13cm 7.5cm},clip,width=.24\textwidth]{XX-graphics/REML.pdf}}
    \subfigure[Cat 4: Retrieval with memory \& feedback]{\label{fig:reml:d}        \includegraphics[trim={13.5cm 0cm 3.5cm 7.5cm},clip,width=.24\textwidth]{XX-graphics/REML.pdf}}
    \vspace{-0.4cm}
    \caption{Retrieval-enhanced machine learning models should implement three necessary requirements (querying, retrieval, and response utilization) and may implement two optional properties (storing information and providing feedback to the information access model). This results in four categories of REML models presented above.}
    \vspace{-0.3cm}
    \label{fig:reml}
\end{figure*}

\section{Motivation}
\label{sec:motivation}

Despite the success of modern high capacity models, focusing on the number of parameters as a primary mechanism to improve performance can be brittle, unsustainable, and opaque \cite{bender:parrots}.  
We argue that these concerns can be addressed by developing ML models that, instead of encoding knowledge in parameters, can access large collections of information items using efficient, effective, and robust retrieval technologies.  
Some of the major applications of REML is presented below:

\noindent \textbf{Generalization.} Recent work has shown that many ML models can significantly benefit from simple retrieval augmentation approaches. For instance, KNN-LM~\cite{khandelwal:knnlm} linearly interpolates large language model predictions with the nearest neighbors of the given context input. This approach does not even require further training or fine-tuning. The authors showed substantial improvements in terms of language model perplexity in both in-distribution and out-of-distribution test sets, demonstrating the generalizability of this approach. KNN-LM together with several other examples reviewed in Section~\ref{sec:case_studies} suggest that enhancing ML models using retrieval models will have a large impact on the generalizability of the models. Retrieval enhancement is expected to have large impact on domain adaptation, zero-shot, and few-shot learning tasks. 

\noindent \textbf{Scalability.} ML models compress information from training data to support accurate prediction at inference time.  Although increasing model capacity by adding parameters often translates into an improvement in predictive power, recent studies demonstrate that large deep learning models often memorize training instances and concepts associated with them in their model parameters \cite{carlini21extracting}. As an alternative to such implicit memorization, retrieval systems can explicitly store information either directly from the training set or from  concepts derived during the learning process.  Because retrieval architectures are often designed to scale, a retrieval system can provide efficient access to this information, substantially reducing the need for high capacity models and increasing throughput.

\noindent \textbf{Collection Updates and the Temporal Aspect.} Current ML models make predictions solely based on the data observed during training.  Although effective in stationary domains, this approach can be brittle in nonstationary domains, such as news, where new information constantly emerges.  And, while periodic retraining is possible in some slowly-changing domains, for quickly-changing domains, this solution is impractical.  An information access system can decouple reasoning from knowledge, allowing it to be maintained and updated independent of model parameters at a cadence aligned with the corpus.

\noindent \textbf{Interpretability and Explainability.} Because the knowledge in training data is encoded in learned model parameters, explanations of model predictions often appeal to abstract and difficult-to-interpret distributed representations.  By grounding inference on retrieved information, predictions can more easily be traced specific data, often stored in a human-readable format such as text.

\noindent \textbf{On-Device Machine Learning.} State-of-the-art ML models require significant computational power and memory availability, which are not available on devices such as smartphones. Retrieval-enhanced ML models can potentially decouple  memorization from generalization and store a large collection (memory) of information items on a remote server. Thus, a small, efficient ML model can be hosted on-device. By minimizing the interactions between the retrieval component and the ML model, this  can potentially revolutionize the applications of on-device machine learning. If privacy is an issue, the information items stored on the remote server can be encrypted and methods, such as the recently developed distance-preserving encryption schemes for nearest neighbor search \cite{Fuchsbauer2021PrivateANN}, can be adopted for privacy-preserving retrieval.


Collectively, these properties of IR techniques suggest the development of REML, which we pursue in the subsequent sections.




\section{Retrieval-Enhanced Machine Learning}
\label{sec:framework}

This paper focuses on predictive ML models.  Let $\mathcal{X}$ be the input (feature) space for the task and $\mathcal{Y}$ be the output (prediction) space. Given an input $x \in \mathcal{X}$, a ML model produces a prediction in the output space $\widehat{y} \in \mathcal{Y}$. Supervised learning models are often trained by minimizing an empirical prediction loss (error) over instances in a training set $T = \{(x, y) \in \mathcal{X} \times \mathcal{Y}\}$.

\textit{Retrieval-enhanced machine learning} (REML) refers to models composed of two coupled components: one model that makes predictions by communicating with $N$ models each  mediating access to a repository of information or knowledge. A REML model is defined as $f_{\theta}(x; R_{\omega_1}, R_{\omega_2}, \cdots, R_{\omega_N})$. The model $f_\theta$ parameterized by $\theta$ is called the \emph{prediction model} and $R_{\omega_i}$ denotes the $i$\textsuperscript{th} \textit{information access model} parameterized by $\omega_i$. Thus, to produce $\widehat{y}$, the prediction model can interface with $N$ information access models. Each $R_{\omega_i}$ includes a collection or repository $C_i$ that is available---through an information access model---to the prediction model. This repository could be composed of natural language documents---as with text retrieval---or some other indexed representation.  As such, $C_i$s reflect a large set of parameters available to the model that can be leveraged \textit{ad hoc}, as with many non-parametric and lazy learning techniques. 
The goal of retrieval-enhanced supervised learning models is to minimize the empirical risk,
\begin{equation}
    \frac{1}{|T|} \sum_{(x, y) \in T}{\mathcal{L}\left(f_{\theta}(x; R_{\omega_1}, R_{\omega_2}, \cdots, R_{\omega_N}), y\right)}\label{eq:objective}
\end{equation}
where $\mathcal{L}$ is a loss function for each training instance.

\subsection{Overview}
We define the following \textit{necessary} requirements (Reqs) for REML:
\begin{enumerate}
    \item[Req 1] \textbf{Querying}: the prediction model $f_{\theta}$ should be able to submit input-dependent queries to the information access models, i.e., $R_{\omega_i}$s.
    \item[Req 2] \textbf{Retrieval}: each information access model $R_{\omega_i}$ should be able to efficiently process the prediction model's queries and retrieve relevant information items from a memory or collection $C_i$.
    \item[Req 3] \textbf{Response Utilization}: the prediction model $f_{\theta}$ should utilize the response returned by the information access models for making predictions.
\end{enumerate}

\begin{figure}[t]
    \centering
    \includegraphics[trim={2cm 1.5cm 3cm 1.5cm},clip,width=\linewidth]{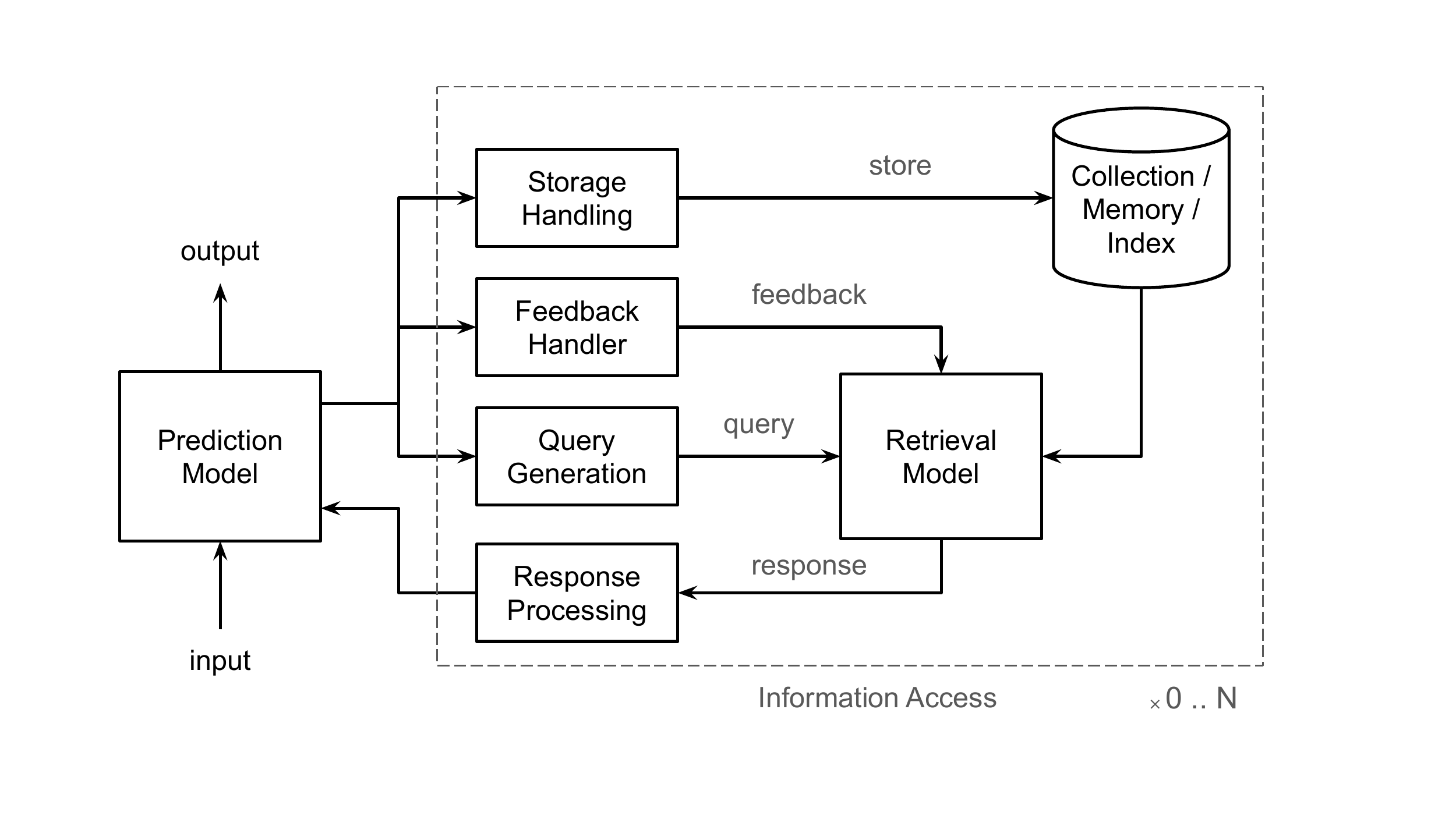}
    \vspace{-0.6cm}
    \caption{A generic framework for REML.}
    \vspace{-0.5cm}
    \label{fig:reml-framework}
\end{figure}

Considering these three requirements, we can envision the first category of REML models. A high-level overview of models in this category is presented in \figurename~\ref{fig:reml:a}. Most existing retrieval-enhanced ML models, such as REALM~\cite{Guu+al:2020}, belong to this category.

REML may also benefit from two additional \textit{optional} properties:
\begin{enumerate}
    \item[Opt 1] \textbf{Storing}: the prediction model may store some information items in a memory for future access during both training and inference. Such information items will be accessible to the model through querying (Req 1).
    
    \item[Opt 2] \textbf{Feedback}: the prediction model may be able to provide feedback to the information access models. This enables the information access models to improve based on the feedback.
\end{enumerate}

\figurename~\ref{fig:reml:b} depicts the second category of REML models that take advantage of Opt 1 by storing information in a memory and accessing the information later. On the other hand, \figurename~\ref{fig:reml:c} demonstrates a high-level overview of the third category of REML models that can provide feedback (Opt 2) to the information access systems. The last category (\figurename~\ref{fig:reml:d}) implements both of these optional properties and supports querying, utilizing retrieval responses, storing information, and providing feedback to the information access systems.


Based on these requirements and optional properties, \figurename~\ref{fig:reml-framework} envisions a generic framework for REML. The framework consists of two major parts: the prediction model $f_\theta$ and the information access models $R_{\omega_i}$s. For each input $x$, the model $f_\theta$ may decide to run multiple retrieval processes by either submitting multiple queries, accessing multiple data repositories and/or memories, providing feedback to the information access component, or a combination of the above. The number of retrieval processes can be zero for some inputs, and thus REML generalizes typical predictive modeling.

\subsection{Information Access in REML}
\label{sec:retrieval}
In its most generic form, each information access system in the proposed REML framework consists of five components: (1) Query Generation, (2) Retrieval Model, (3) Response Processing, (4) Feedback Handler, and (5) Storage Handler. In the following subsections, we discuss potential implementations for each component.

\subsubsection{Query Generation}
In current information access systems, queries mostly take the form of unstructured text (e.g., keyword queries or natural language questions), structured query language (e.g., SQL), or multi-media items (e.g., images). Such query languages and formats can be also adopted by retrieval-enhanced ML models. The Query Generation component is responsible for generating one of these query formats. Note that depending on the application and due to efficiency or effectiveness requirements, one may simply cast the query generation problem to query selection from a set of pre-defined queries. In either case, the Query Generation (or Selection) component should be able to translate the information need of the prediction model $f_\theta$ to a query language or format that can be efficiently processed by the information access model $R_{\omega_i}$. Since retrieval models accessible by $f_\theta$ may accept different query languages, the Query Generation component may be unique to each retrieval model.

Existing information access systems are designed for people and, therefore, existing query formats (mentioned above) are understandable by people. In the context of REML, we can relax the requirement of an interpretable query language. Besides the common query languages and formats, the prediction models can produce any \textit{latent representation} (e.g., a high-dimensional dense vector) as a query. For instance, any hidden layer representation produced by the prediction model $f_\theta$ may be used as a query for retrieval. That being said, queries may also be generated from the input $x$ itself without any involvement of the prediction model parameters. 

Under REML, prediction models do not have restrictions on the number of queries that can be submitted for each input $x$. As a result, a model may generate multiple, sequential queries produced for each input $x$, resulting in a \textit{query session} analogous to human search sessions. While current search engines base sessions on temporally-adjacent user queries, REML prediction models can, when querying, explicitly indicate a unique session ID associated with the input $x$.

\subsubsection{Retrieval Model}
The retrieval model component aims at retrieving information items from the given collection, repository, or memory in response to each query produced by the Query Generation component. Existing retrieval models are mostly designed based on the probability ranking principle (PRP) \cite{Robertson1997PRP}, in which documents are ranked based on their probability of relevance to the query. In the IR literature, relevance can be defined in five levels \cite{Saracevic1996relevance}: (1) systematic or algorithmic relevance, (2) topical relevance, (3) cognitive relevance or pertinence, (4) situational relevance or utility, and (5) motivational or affective relevance. 
However, these definitions assume that the retrieved documents are consumed by humans. This assumption no longer holds for REML models, thus the notion of relevance needs to be revisited for REML. 

When designing retrieval models for REML,  relevance can be thought of as the utility that the prediction model obtains by consuming the results produced by the retrieval model; this is similar to task-based perspectives on (human) information retrieval \cite{kelly:task-workshop}.
For simplicity and without loss of generality, assume for each input $x$, the prediction model $f_{\theta}$ only submits a single query $q$ to a retrieval model that returns a result list $L_q = \{(d_1, \phi(d_1)), (d_2, \phi(d_2)), \cdots,$ $(d_k, \phi(d_k))\}$, where each $d_i$ is a document\footnote{In this paper, we refer to retrievable items, e.g., unstructured text, image, or even latent vectors, as documents.}  in the collection and $\phi(d_i)$ encodes a list of features and properties associated with document $d_i$. For instance, $\phi(d_i)$ may contain the document score produced by the retrieval model in addition to a number of features used by the retrieval model to compute the score. With a slight abuse of notation, let $f(x; L_q)$ denote the prediction function that submits the query $q$ to a retrieval model and uses its response (i.e., $L_q$) to make a prediction. Then, the utility gain can be defined as:
\begin{equation}
    \text{UtilityGain}(q, L_q; f_{\theta}, x) = U(f_{\theta}(x; L_q), y) - U(f_{\theta}(x; \emptyset), y)
    \label{eq:rel}
\end{equation}
where $U(\cdot, \cdot)$ represents some desired utility function. This definition assumes that data points $(x, y)$ are i.i.d. samples. Utility gain depends on how the prediction model $f_\theta$ consumes $L_q$ for producing $\widehat{y}$. Utility gain can take on both positive and negative values. A negative gain means that the retrieval results $L_q$ have negative impact on predicting the ground truth label. This definition can be extended to multiple queries per $x$. 

The implementation of retrieval models for REML depends on the nature of documents in the collection. For instance, one can use the vector space model and employ the inner product as the similarity function between query and document vectors. Section~\ref{sec:optimization} provides more information on the optimization of retrieval models in REML.

\subsubsection{Response Processing}
The way the prediction models consume the retrieved items has a substantial impact on their end-to-end performance. The Response Processing component takes the results returned by the retrieval models for each query $q$ (i.e., $L_q$s) and prepares it for consumption by the prediction model. 

This component can be implemented by returning the content of the retrieved documents, synthesizing a summary of their content, producing one or more semantic representations of their content, combining all the information presented in $L_q$ in some way, and so on. There are many design choices here and the best choice will largely depend on the nature of the machine learning model and the task it is being applied to.

\subsubsection{Feedback Handler}
When training  retrieval models, it is often desirable to get feedback from the machine learning model. Such feedback can then be used as a signal for optimizing the retrieval model. We can imagine various forms of feedback in this context. For example, the model can compute the utility gain of documents returned by the retrieval model using Equation~\eqref{eq:rel}. As another example, the feedback may be computed based on the gradients of the prediction loss with respect to the retrieved information. Section~\ref{sec:optimization} discusses how the model's feedback can be used for optimizing retrieval models in REML.

\subsubsection{Storage Handler}
If the prediction model has the ability to store information in the repository (or memory), the Storage Handler can expand the collection by storing the information item into the memory. 
However, for efficient storage and access of a large number of items, careful consideration of memory management techniques, hardware requirements, and storage data structures beyond existing technologies (e.g., inverted indexes) is required. Besides information storage, this component is also responsible for storage management. Thus, it should implement caching, compression, access controls, and time-to-live requirements as necessary.


\subsection{REML Optimization}
\label{sec:optimization}
We envision three optimization approaches for REML: (1) independent optimization of prediction and information access models, (2) conditional optimization of these models such that the quality of one impacts the optimization of the other, and (3) joint end-to-end optimization of both models. Without loss of generality, here we assume that there only exists one information access model.


\subsubsection{Independent Optimization of Prediction and Information Access Models}
In independent optimization, the training process of the prediction model $f_\theta$ is independent of the retrieval performance. For example, we can assume that the retrieval model is optimal. Formally, we can optimize the prediction model of REML as:
\begin{align}
    \theta^* = \arg \min_{\theta} \frac{1}{|T|} \sum_{(x, y) \in T} \mathcal{L}(f_{\theta}(x; R_{\text{opt}}), y)
\end{align}
where $R_{\text{opt}}$ denotes an optimal retrieval model and can be modeled using ground truth relevance information, if available. Similar to~\cite{Zamani:2018:TheoryWS}, we can also model imperfect retrieval models by introducing noise to an optimal ranking behavior. The retrieval model can be trained using typical learning-to-rank (LTR) formulation, independent of $f_\theta$. For the same of space, we refer the reader to \citet{Liu2009LTR} for more information on LTR models.

\subsubsection{Conditional Optimization of Prediction and Information Access Models}
In conditional optimization, the prediction model parameters get updated conditioned on the retrieval model's performance and vice versa. This process can be done iteratively until a stopping criterion is met (e.g., convergence or early stopping based on performance on a held-out validation set). Therefore, the prediction model can be optimized as:
\begin{align}
    \theta^{(t)} = \arg \min_{\theta} \frac{1}{|T|} \sum_{(x, y) \in T} \mathcal{L}(f_{\theta}(x; R_{\omega^{(t)}}), y) \\
    \omega^{(t+1)} = \arg \min_{\omega} \frac{1}{|T|} \sum_{(x, y) \in T} \mathcal{L}(f_{\theta^{(t)}}(x; R_{\omega}), y) 
\end{align}
where $\theta^{(t)}$ and $\omega^{(t)}$ denote the parameters of the prediction model and the information access model at the $t$\textsuperscript{th} iteration, respectively. These equations assume that both models are being optimized. In case of using unsupervised retrieval models, the second optimization process would be skipped (i.e., $\omega_{t+1} = \omega_t$).


\subsubsection{Joint End-to-End Optimization}
In end-to-end optimization of REML, both ML and information access models are trained jointly by optimizing a single objective function. Formally, it is defined as:
\begin{align}
    \theta^*, \omega^* = \arg \min_{\theta, \omega} \frac{1}{|T|} \sum_{(x, y) \in T} \mathcal{L}(f_{\theta}(x; R_{\omega}), y) 
\end{align}

For optimizing this objective via gradient descent-based optimizers, the whole REML process (both models and their interactions) is required to be differentiable. End-to-end optimization is expected to perform better than the last two optimization approaches, but given the complexity of retrieval from large collections, this requirement may be difficult to satisfy in some cases.

\subsection{Extending REML to Multiple ML Models}
\label{sec:multiple_models}
Previous sections consider only a single prediction model that interacts with multiple retrieval processes  (see \figurename~\ref{fig:reml-framework}). This section extends the REML framework to multiple prediction models. Similar to current search engines that provide service to many users, retrieval models can be also employed by multiple ML models. 

Assume there are $M$ prediction models $f_{\theta_1}, f_{\theta_2}, \cdots, f_{\theta_M}$ that use $N$ information access models denoted by $R_{\omega_1}, R_{\omega_2}, \cdots, R_{\omega_N}$. Each $R_{\omega_i}$ should provide service to multiple prediction models. This introduces the following challenges:

    \noindent \textbf{Shared Query Language}: All prediction models may need to share the same query language for interacting with retrieval systems. 
    
    \noindent \textbf{Shared Response Formats}: The responses produced by each retrieval system will be used by all prediction models. Therefore, the prediction models should be able to utilize the response format used by each retrieval model.
    
    \noindent \textbf{Shared Storage}: The storage used by each retrieval model is shared between all prediction models. Storage is a limited resource, thus a policy may be required to regulate storage usage for each prediction model. Moreover, the data stored by each prediction model may not be interpretable by other models or may not be shared due to privacy restrictions. The Storage Handling component should develop memory management and access restriction policies and functionalities for each storage request. 
    
    \noindent \textbf{Personalization}:\footnote{Personalization is often used for humans. We stick to the same terminology to be consistent with the IR literature.} The prediction models have special needs and they utilize the retrieval responses differently. Therefore, in response to a query $q$ submitted by two prediction models $f_{\theta_i}$ and $f_{\theta_j}$, the retrieval models may want to respond differently. In this case, retrieval models would need to implement models and techniques for personalizing the search results. 
    
    \noindent \textbf{Comparable Feedback Across Prediction Models}: Comparable feedback across prediction models enables us to easily aggregate the obtained feedback. Otherwise, the feedback can be used for each individual prediction model as a form of personalization. 
    
    \noindent \textbf{Optimizing Retrieval Models}: In case of dealing with trainable retrieval models, the optimization solutions introduced in Section~\ref{sec:optimization} need further adjustments. Let $\mathcal{L}_i$ denote the loss function associated with the $i$\textsuperscript{th} prediction model. Thus, the joint end-to-end optimization of models can be achieved as follows:
    \begin{align}
        \arg \min_{\theta, \omega} \frac{1}{M} \sum_{i=1}^{M}
        \frac{1}{|T_i|} \sum_{(x, y) \in T_i} \alpha_i \mathcal{L}_i(f_{\theta_i}(x; R_{\omega}), y) 
    \end{align}
    where $T_i$ denotes the training data for the $i$\textsuperscript{th} prediction task. This formulation assumes that the loss values are comparable across prediction models. The hyper-parameter $\alpha_i$s control the weight of each loss function. The conditional optimization formulation can be adjusted, similarly. 


\subsection{Information Access Evaluation in REML}
\label{sec:eval}
The prediction model should be evaluated based on its performance on the downstream task, and appropriate evaluation methodologies and metrics should be chosen considering the downstream task. This evaluation is the same for any predictive model designed for that task. Therefore, we skip the evaluation of prediction models and discuss approaches for evaluating the information access models. Evaluating information access in REML is particularly important for diagnosing the retrieval process and designing retrieval systems that provide service to multiple prediction models (see Section~\ref{sec:multiple_models}). The retrieval component in REML can be evaluated either extrinsically or intrinsically:

\noindent \textbf{Extrinsic Evaluation:}  The information access quality can be quantified by measuring its impact on the prediction model for the downstream task. This is perhaps the most important factor in evaluating information access in REML. Note that in case of having multiple prediction models, extrinsic evaluation is defined for each prediction model independently. However, aggregating the downstream performances for different prediction models is challenging, because prediction models may be evaluated based on various metrics and methodologies and they may not aggregate easily.
Extrinsic evaluation can be done both through offline and online evaluation. 

\noindent \textbf{Intrinsic Evaluation:} In intrinsic evaluation, the retrieval model is evaluated independent of the prediction models. To do so, one may define \textit{relevance} based on the desired documents expected to be retrieved for a prediction model. This definition may be obtained from experts or by analyzing observations from prediction models' behavior. Then presumably an annotation process, e.g., through pooling, may be employed for creating data collections for intrinsic evaluation of the information access model. Metrics used in intrinsic evaluation are expected to have high correlations with the downstream performance of the prediction models. We highlight that most metrics used in the IR literature have been developed based on user behaviors with search engines. For instance, many of them assume that users assess documents sequentially. However, such assumptions may not hold for many ML models. Thus, new evaluation metrics may need to be developed.


\section{Case Studies}
\label{sec:case_studies}
Since REML is a general framework, we can discuss related approaches as special cases of REML.  This exercise helps us understand how and when REML might work and suggests opportunities for extending existing work. 

\subsection{Knowledge Grounding}
Fully data-driven ML models, despite demonstrating success across a wide number of tasks, still lack grounding in the real world. Access to external knowledge, via \textit{knowledge grounding}, may help with this issue~\citep{zhu2021retrieving,dehghani2019learning,komeili:blenderbot2,Lewis+al:2020,Hashemi2020GT}. Knowledge grounding models make predictions based on the results returned by a retrieval model. 

In the context of language modeling, one class of methods uses retrieval results as evidence to support \textit{reasoning}. For example, the knowledge retriever module in REALM~\citep{Guu+al:2020}  accesses information from an encoded Wikipedia corpus during pre-training. In text generation, RetGen~\citep{zhangAAAI22} combines a grounded text generator with a document retriever. Grounding the generation helps with the issue of hallucinated facts, and the retrieval component makes the grounding effective and efficient. 
\citet{Lewis+al:2020} highlighted the importance of retrieval in knowledge-intensive NLP tasks and introduced retrieval-augmented generation (RAG) by augmenting a generator with the output of a non-parametric retriever that uses maximum inner product search.

Entities as Experts (EaE)~\citep{fevry-etal-2020-entities} introduces an entity memory that can be accessed by the model and the retrieved representations of entities are combined with the input representation for entity linking, mention detection, and masked language modeling tasks. Similarly, Fact as Experts (FaE)~\citep{Verga+al:2021} incorporates a fact memory for language modeling. Such a mechanism gives access to factual information, that may expand or change over time, while there is no need for additional training or fine-tuning.

In open-domain QA, a common approach is to retrieve documents or passages from Wikipedia or even the Web and then extract answers~\cite{izacard2020leveraging,Qu2021OKVQA}. \citet{lee2019latent} used an encoded Wikipedia corpus to train a retrieval model and then fine-tune the prediction model for a QA objective. \citet{khattabNEURIPS21} used a retrieval component for multi-hop reasoning, where the retrieved facts from each hop are summarized into a short context and becomes a part of the query for the subsequent hops. Similarly, \citet{das-etal-2019-multi} performed iterative retrieval for expanding and rewriting multi-hop questions. 
This is also the case for task-oriented dialogues. 
For instance, LaMDA~\citep{Thoppilan:2022} shows the benefit of granting dialogue systems access to external knowledge for reducing  unsourced statement hallucination~\citep{shuster2021retrieval}. 

The approaches presented in this subsection mostly use simple retrieval models, e.g., TF-IDF or inner product of learned representations, for finding factual information from external knowledge bases. Therefore, one can look at knowledge grounding as an implementation of REML, mostly based on Category 1: Retrieval-only (\figurename~\ref{fig:reml:a}) or Category 3: Retrieval with feedback (\figurename~\ref{fig:reml:c}). 

\subsection{Memory-Augmented Machine Learning}
Using a memory component where the model can read from and/or write into is one of the most common ways of implementing REML in neural networks. 
The main motivation is to use an explicit storage buffer to make it easier for the network to rapidly incorporate new information and not to forget in the future. 

A model may use an internal memory where it compresses and accumulates information to access them in later stages of the process. This has been the base of several neural architecture classes. For instance, Long Short-Term Memory networks (LSTMs)~\citep{hochreiter1997long} or Gated Recurrent Networks~\citep{cho2014learning} that use a latent state as a memory to collect information from previous time steps. Attention-based models~\citep{bahdanau2014neural, vaswani:transformer} also treat different parts of the input as memories and use soft access as the retrieval mechanism to manage the interaction between them.
However, memory-augmented neural networks refers to cases of using an external memory~\citep{santoro2016meta}. Among main works in this area, memory networks~\citep{sukhbaatar2015end} explicitly store information in a form that is element-wise addressable. 
Neural Turing machines~\citep{graves2014neural, graves2016hybrid} are well-known examples of ML models that can read from and write into an external memory matrix in order to represent and manipulate complex data structures.

The common target property of memory-augmented neural networks is incorporating an external memory that is trained end-to-end with the objective and data from downstream tasks. This most resonates with the fourth category of REML: Retrieval with memory and feedback (\figurename~\ref{fig:reml:d}). However, the memory size in existing models is relatively small and extending the memory size is an exciting and challenging research direction.

\subsection{Retrieval-Enhanced Input Representation}
A number of retrieval-enhanced models use the retrieved items to update the representations of the model's input. This is different from knowledge grounding in the sense that the information items do not necessary include the knowledge required for accomplishing the task. Instead, the retrieved information contains patterns that can help the model to learn more expressive representations. 

Pseudo relevance feedback (PRF) is an example of such models. It uses the top retrieved documents for updating the query representation through query expansion. It has shown successful results in a wide range of retrieval tasks \cite{Attar:1978,Croft:1979,Lavrenko2001RM,Zhai2001Mix,Xu:1996,jeonSIGIR03,Feng2004}, demonstrating the quality of the produced query representations for retrieval. 
%
Recently, \citet{Hashemi2020GT} proposed Guided Transformer, an extension to the Transformer network that includes cross attention for contextualizing inputs with retrieved information from multiple information sources to learn more accurate representations of the model's input. In their subsequent work \cite{Hashemi2021NMIR}, the authors proposed an approach for learning multiple representations for query intents by utilizing the retrieval results and taking advantage of the Guided Transformer network for representation adjustment. More recently, \citet{borgeaud:retro} proposed RETRO for language modeling and showed that by using networks like  Guided Transformer one can enable access to a trillion-scale database for a relatively small model. 

Related approaches have been also used in computer vision \cite{Gur+al:2021,kuo2020mask2cad,li2015database,siddiqui2021retrievalfuse}. For example, \citet{xu2021texture} studied the task of image inpainting whose goal is to restore missing regions of an image. They introduced a ``texture memory'' that augments a neural network with access to patches extracted from unmasked regions of the input image.
For the task of 3D scene reconstruction, \citet{siddiqui2021retrievalfuse} used retrieval for creating multiple approximate reconstructions and then fusing them with an attention-based blending module to generate the output.  
For object detection, \citet{kuo2020mask2cad} used retrieval from a large-scale dataset of 3D models to understand the underlying 3D structure of objects seen in a 2D image.

Similar to knowledge grounding, retrieval-enhanced representation learning can take advantage of information items that are similar to the input by learning from patterns observed in the retrieved results. Thus, the first (retrieval-only) and the third (retrieval with feedback) REML categories are often used for this purpose. 



\subsection{Generalization through Memorization}
Combining retrieval-based and generative approaches has been explored in a number of applications. In this case, the retrieval component can contribute by producing accurate responses when memorization is sufficient.

Motivated by the goal of memorizing rare patterns explicitly, \citet{khandelwal:knnlm} introduced KNN-LM, where a retrieval mechanism is used to find the nearest neighbor tokens given the prefix as query. KNN-LM linearly interpolates the predicted distribution for the next token using distance information from the retrieval mechanism. BERT-KNN~\citep{kassner-schutze-2020-bert} employs a similar nearest neighbor algorithm to augment a BERT model to learn better representations for rare facts. This idea has also been extended to machine translation~\cite{khandelwal2021nearest}. It is shown that retrieval augmentation improves domain adaptation by using a domain-specific datastore for retrieval. \citet{tay2022transformer} proposed training a large model that memorizes the mapping of document content to document ids, which can be used to retrieve relevant document ids given a query at inference time. This model could be an alternative to KNN based models we discussed above to serve a REML system as a differential index.

In dialogue systems, given a dialogue history as a query, a retrieval unit can be used to return the top ranked candidate response as the next dialogue utterance~\citep{roller:blenderbot}. Such retrieval-based approaches can also be combined with response generation models and form a hybrid solution for dialogue systems~\cite{Yang2019RetrGenConv}.

Another approach to improve generalization through memorization is through updating retrieval results. In some cases, editing an existing candidate output is easier than generating it from scratch, especially in complex structured output generation tasks, like code generation. \citet{Hashimoto:2018} proposed to retrieve a training example given the input and edit it to the desired output. The retriever and the editing modules are trained jointly. 
\citet{Pasupat+al:2021} proposed using exemplar retrieval for semantic parsing. In their setup, given a query, the parser retrieves a set of related exemplars, augments the query using the retrieved information, and then incorporates a seq2seq model~\cite{Sutskever:2014:seq2seq} to produce an output parse. 

The aforementioned methods try to use a retrieval component to handle memorization cases. It is found useful, especially for cases where sufficient training data is not available. Many existing models are based on a retrieval-only implementation of REML. 

\subsection{Efficient Access to Longer Context}
Due to memory constraints as well as efficiency and effectiveness reservations, consuming and representing large inputs, e.g., long text documents or videos, are challenging. 
REML offers a solution to address this issue by giving access to the context of any size via a retrieval mechanism. Here we mention a few examples of studies that exploit this idea.


\citet{wu2019long} proposed using a long-term feature bank for detailed video understanding. The long-term feature bank stores a rich, time-indexed representation of a long video. Then the video understanding model consults with the bank through a retrieval module to get features that encode information about past and future scenes, objects, and actions. 
Similarly, MemViT~\citep{wu2022memvit}, proposes to process videos in an online fashion and store information in memory at each iteration. The model can retrieve prior context from the memory to enable long-term modeling for the recognition task.
Similar approaches have also been used for video object segmentation~\cite{oh2019video} and video summarization~\cite{lee2018memory}. 
%

For processing long documents, researchers often split the documents into passages. For instance, \citet{Dai2019} only used the first passage of each document for document retrieval. \citet{Xiong2021ANCE} used the passage with the maximum similarity score with the query. The end-to-end intra-document cascading model \cite{Hofstatter2021ICDM}  can be seen as a REML model with feedback. It first selects (retrieves) a number of passages from the document and then consumes the selected passages for scoring the document.

The methods presented in this subsection are perhaps the simplest implementations of REML: the retrieval collection is not large, and some of them do not use feedback. 

\subsection{Retrieval-Enhanced Optimization}
All the methods mentioned above use a retrieval component at the inference time for making accurate predictions. Some approaches use retrieval components solely for the purpose of optimization, e.g., for producing training data and/or computing loss functions. Thus, the retrieval model will not be used during inference. 

A natural application of retrieval-enhanced optimization is for retrieval tasks. \citet{Dehghani2017Weak} introduced a weak supervision approach for IR by producing large-scale training data through BM25 and training ML models for document ranking. 
\citet{Zamani2017RelWE} used the top retrieved documents to produce a relevance model distribution for training queries and learn relevance-based word embedding. 
Producing hard negative instances for training learning-to-rank models is another application of REML. For instance, ANCE~\cite{Xiong2021ANCE} and its extensions \cite{Prakash2021RANCE,Li2021DANCE} are dense retrieval models that iteratively use the model parameters to retrieve documents for producing `hard' negative samples for training the model. 

\citet{Wu+al:2019} used a retrieval unit to enable unsupervised training of machine translations, i.e., using two monolingual corpora in the source and target languages with no alignment. As an alternative to back translation, they proposed retrieving a sentence from the target corpus using the source sentence and applying some changes using an editing mechanism to the retrieved target to generate source-target pairs and train the MT model.
\citet{Triantafillou2017FewShotIRLens} proposed an approach for few-shot learning through retrieval. This approach retrieves items for each input and uses them for making predictions. 
Via this approach, a model can adapt to a new domain without additional training or new data.

An interesting use case of REML is the pre-training task of CLIP~\citep{radford2021learning} and VideoCLIP~\cite{xu-etal-2021-videoclip} which are practically optimizing for text-image and text-video retrieval, respectively. They are in fact capturing cross-modal relevance that led to learning representations that are effective in various setups, like zero-shot classification.   

\section{A Research Agenda}
\label{sec:research-agenda}
While Section~\ref{sec:case_studies} provides evidence of the efficacy and broad applicability of REML, there remain significant open research challenges in fully realizing the general REML vision, some of which are already mentioned in previous sections. 

\subsection{Querying}
In developing a prediction model that supports retrieval, understanding how to query becomes a core research question.  First, this involves knowing when to query.  In some situations, a prediction model may not benefit from a retrieval operation (even if it benefits on average).  Although current retrieval-enhanced systems issue the equivalent of queries for \textit{every instance}, when querying incurs some cost, be it in the form of latency or financial expense, developing models that ``know when they don't know'' would allow the prediction algorithm to explicitly trade off cost and benefit.    A prediction model that has access to multiple information access services can make this decision for individual corpora, perhaps select the appropriate source for the instance.  Second, at a more granular level, how retrieval might benefit a model may vary by instance $x$.  For example, retrieval may support uncertainty in one part of the $\theta$ for one instance and uncertainty in another part of $\theta$ for another instance.  This self-interrogation can be explicitly designed or implicitly learned.  Nevertheless, even learnable behavior requires an architecture and parameters to adapt.  
Finally, many retrieval situations can benefit from the searcher conveying non-semantic meta-retrieval information such as uncertainty in (aspects of) the query or context of the retrieval itself.  People often convey similar information to human intermediaries \cite{tot-retrieval} and we suspect that more expressive querying can also emerge in REML.

In developing an information access model to support a prediction model, similar questions arise.  First, developing or learning a query language requires expressiveness that captures the breadth of model needs.  At the same time, it should allow for communication of meta-retrieval or structured properties of the retrieved set.  Moreover, these properties need to be explored within the effectiveness and efficiency constraints.  Second, although a query may be effective and efficient in general, it may be ambiguous or imprecise for a particular retrieval scenario.  This is especially likely in situations where  multiple models may develop inconsistent uses of the query language (Section \ref{sec:multiple_models}).


\subsection{Storing}
\label{sec:research-agenda:storing}
The ability of the prediction model to store items presents unique problems not encountered in traditional retrieval or ML research.  Although architectures like memory networks \cite{weston:memory-networks} provide modestly sized storage, we anticipate models storing or serializing on a larger scale with more permanence.  In situations with multiple models (Section \ref{sec:multiple_models}), we anticipate the corpus operating as a means to share derived knowledge (to avoid re-computation during inference) or prediction model parameters (to support learning).  

In developing a prediction model that supports storage, understanding how to store becomes a core research question.   Just as with querying, a model needs to reason about when to store, what to store from its parameters or reasoning, and how to represent that information.  Each of these questions is relevant both to sharing derived knowledge as well as model parameters.  Like queries, stored items may include auxiliary information such as the model's confidence in the derived data or parameter values, the prediction task, and other information that may be valuable for an information access system to make retrieval decisions.   More so than with queries, a model might need to be more judicious in storage operations, since injecting irrelevant or erroneous content into the corpus can significantly degrade its usefulness.

In developing an information access model to support storage, classic problems related to indexing arise.  First, as with queries, the language, schema, or representation of an item requires careful construction to optimize for effectiveness and efficiency.  Second, in accepting a storage request from a prediction model, the information access system needs to model the value of the content.  Redundant items can either add noise or improve coverage, depending on the task.  Or, an item may require processing to make indexing and retrieval more effective.  These decisions can be based on the content of the item or meta-data about the item, such as the confidence of the model or, in the case of multiple models, confidence in the prediction model itself.  Third, if an item \textit{should} be stored, there is the question of \textit{how} to store it.  This includes questions of item compression and representation, both of which need to occur incrementally but improve  with batch, corpus-wide computation.  Finally, in the case of limited capacity in the retrieval index, storage operations may necessitate purging less effective content.  This requires that the information access model reason about how collection management decisions impact  prediction models.

\subsection{Searching}
Ranking functions, a fundamental property of traditional information access systems, influence design decisions about how to store content compactly, how to search that content quickly, and how to return results effectively.  In moving toward REML, several fundamental research questions need to be addressed in order to satisfy these properties for machines.  First, items in REML indexes are likely to be differently structured than existing text documents (see Section \ref{sec:research-agenda:storing}).  Although representations like dense, fixed dimensional vectors are amenable to efficient storage and retrieval, structures that include uncertainty and other attributes may require embedding as a representation amenable to fast retrieval (e.g., vectors) or different indexing schemes altogether.  Second, the representations of items in the index themselves should be selected for effectiveness in supporting prediction models, as well as the space and runtime efficiency.  In some cases, this means accurate and fast score computation.  When a retrieval involves more elaborate post-processing before returning results, this may mean decomposing items before indexing (as is often done when retrieving passages, as opposed to documents).  Third, in situations where there are multiple prediction models, the information access system can use the identity of the model in order to `personalize' results for that model.  Similarly, we can interpret the feedback from prediction models based on where it comes from; some models may not provide actionable feedback early in learning, others may be quite reliable, while others yet might be adversarial.  Third, these representations and their associated ranking functions themselves should be tunable given feedback from  prediction models (see Section \ref{sec:research-agenda:feedback}).  Adjustments to representations and model weights should  be sensitive to confidence in the feedback signal in situations where feedback includes a confidence estimate or if the information access model can estimate the reliability of the feedback.

\subsection{Information Presentation \& Consumption}
\label{sec:research-agenda:presentation}
Representing the retrieval results  in traditional information access involves returning a ranked list of items.  Although items include scores, these are often only used to sort items and are rarely presented to the user.  In the context of REML, we can consider more elaborate representations of retrieval results because they are being consumed by machines.  This introduces a number of exciting research directions.  First, system designers will need to understand the appropriate information to communicate to prediction models, be it an item ranking, a scored set, a set where each item is associated with a score distribution, a graph of inter-item relationships, or some other object derived from the retrieval.  Each of these choices needs to satisfy improving the prediction model's effectiveness, within any cost constraints (e.g., bandwidth, compute).  Moreover, in situations with multiple prediction models, the consistency, interpretability, and maintainability of this representation language become extremely important.  Second, from an efficiency perspective, just as computing a top $k$ ranking can suggest fast document scoring, information about the representation can introduce opportunities for more efficient computations of objects like graphs and score distributions.  Third, a prediction model with access to multiple information access models needs to reason over multiple sets of results.  Information  encoded in the results--explicitly or not--can allow the prediction model to consider the reliability of results before incorporating them into inference. Finally, from a machine learning perspective, \textit{how} to incorporate results into inference will become an important area of work.  Current approaches based on neighbors provide a simple approach, although more sophisticated techniques are likely to improve performance.  


\subsection{Feedback}
\label{sec:research-agenda:feedback}
Modern information access systems use implicit user feedback in order to optimize model parameters.  Although we can imagine a prediction model providing loss information in its feedback similar to how users might provide slate-level feedback, machines may be able to convey more granular and expressive feedback to the information access model.   As such, the first area of research centers on forms of feedback, including scalar values, vectors of values, and more expressive data with goal of helping the information access model improve.  While single scalar feedback values seem simplest, even modern search engines exploit implicit item-level feedback.  We can imagine more targeted and attributed feedback provided by the prediction model.  This structured feedback can include attribution to different components of the retrieval structure (Section \ref{sec:research-agenda:presentation}).  Of course, this requires the prediction model being able to identify the relationship between prediction error and different parts of the retrieval result; in the case of multiple information access services, attribution to individual corpus results.  The second area of research focuses on how an information access model might adjust model parameters given rich feedback from the prediction model.  Current ranking models, with appropriate treatment of different biases, can interpret user feedback as attributed to individual items in the ranking.  A machine may be able to provide feedback that has fewer biases and better calibration than human feedback.  This includes exploring a new space of feedback beyond scalar item-level values.  This also calls for novel approaches for optimizing information access models based on the provided feedback.

\subsection{Evaluation}
The objective of REML is to support machines.  As such, standard methods of evaluating modeling performance (e.g., Equation~\ref{eq:objective}) can be adopted to assess prediction model performance.  Nevertheless, REML introduces several research directions around model evaluation.  First, because of the large, flexible storage capacity, REML can memorize training data or cache previous predictions, resulting in performance metrics (e.g., accuracy) conflating a model's ability to reason (i.e., the prediction model) and its ability to remember (i.e., the information access model).  Methods of selecting evaluation instances or ablation experiments can isolate the contribution of each component.  Second, in situations with multiple prediction models, we need methods to assess performance changes for a group of models with a shared information access service.  Although these per-model losses can be aggregated into a simple average, this may obscure model- or task-specific under-performance.  That said, in some situations, storage operations might result in sharing information, boosting collective performance, and necessitating an evaluation method that decouples reasoning from memorization.  Finally, efficiency metrics that capture the cost of query and response operations (e.g., latency, financial) will need to be developed.

In some cases, we are interested in evaluating the information access model in isolation to make a claim about generalizability of a specific retrieval model to new prediction models, just as we traditionally consider evaluation queries as a \textit{sample} from the full set of queries we would like to apply a system to.  Although we can evaluate information access models using the existing information access evaluation methods (e.g., Cranfield-style offline evaluation, click feedback), we anticipate the opportunity---and sometimes \textit{need}---to develop entirely new evaluation schemes.  First, although a prediction model can be evaluated by its loss function, an information access model can be evaluated by its adoption.  Indeed, if a retrieval component is not used, then perhaps it can be removed altogether.  To see why retrieval systems may be more or less valuable over time, consider the situation where a prediction model can store items such as partial inference or complete inference; in this case, the storage can act like a cache, with queries likely to grow with time, depending on the data.  Or, if there are multiple information access services, the usefulness of some may increase or decrease over time.  Nonstationarity can also arise  when instances have serial dependencies, such as when a retrieval system is repeatedly queried during a dialog or multi-hop task.  Second, estimating an information access model's performance on out of sample domains or tasks requires careful selection of training and evaluation tasks.  Third, in developing offline or batch evaluation methods, although we can avoid some issues, labeling items for relevance and designing metrics reflective of model use becomes difficult, since existing ranking metrics are unlikely to approximate how a machine would consume results (see Section \ref{sec:research-agenda:presentation}).  Finally, REML presents a tremendous opportunity to study these questions \textit{in silico}.  This means that experimentation and analysis, although more complicated, will be much faster than systems serving people, without safety concerns, since experiments can be run isolated from people.


\section{Conclusion}
\label{sec:conclusion}
Although the large number of parameters in models such as deep neural networks has, in part, resulted in impressive improvements in performance across a wide range of tasks, evidence suggests that these successes may be partially due to the increased capacity to store information in model parameters \cite{zhang:noise-memorization}. We claim that, if model capacity is being used to store information, then we should decouple reasoning from memory and expand the scope of information retrieval to also support ML models.  Starting from this claim, we have presented a general framework, its relation to existing methods, and its ability to substantially advance how we think about information retrieval and how we do machine learning.

\section*{Acknowledgments}
This research was supported in part by the Google Visiting Scholar program and in part by the Center for Intelligent Information Retrieval. Any opinions, findings, and conclusions or recommendations expressed in this material are those of the authors and do not necessarily reflect those of the sponsors. We would like to thank Marc Najork for providing feedback on the paper.


\begin{thebibliography}{76}


\ifx \showCODEN    \undefined \def \showCODEN     #1{\unskip}     \fi
\ifx \showDOI      \undefined \def \showDOI       #1{#1}\fi
\ifx \showISBNx    \undefined \def \showISBNx     #1{\unskip}     \fi
\ifx \showISBNxiii \undefined \def \showISBNxiii  #1{\unskip}     \fi
\ifx \showISSN     \undefined \def \showISSN      #1{\unskip}     \fi
\ifx \showLCCN     \undefined \def \showLCCN      #1{\unskip}     \fi
\ifx \shownote     \undefined \def \shownote      #1{#1}          \fi
\ifx \showarticletitle \undefined \def \showarticletitle #1{#1}   \fi
\ifx \showURL      \undefined \def \showURL       {\relax}        \fi
\providecommand\bibfield[2]{#2}
\providecommand\bibinfo[2]{#2}
\providecommand\natexlab[1]{#1}
\providecommand\showeprint[2][]{arXiv:#2}

\bibitem[\protect\citeauthoryear{Arguello, Ferguson, Fine, Mitra, Zamani, and
  Diaz}{Arguello et~al\mbox{.}}{2021}]%
        {tot-retrieval}
\bibfield{author}{\bibinfo{person}{Jaime Arguello}, \bibinfo{person}{Adam
  Ferguson}, \bibinfo{person}{Emery Fine}, \bibinfo{person}{Bhaskar Mitra},
  \bibinfo{person}{Hamed Zamani}, {and} \bibinfo{person}{Fernando Diaz}.}
  \bibinfo{year}{2021}\natexlab{}.
\newblock \showarticletitle{Tip of the Tongue Known-Item Retrieval: A Case
  Study in Movie Identification}. In \bibinfo{booktitle}{\emph{Proceedings of
  the 2021 Conference on Human Information Interaction and Retrieval}}.
  \bibinfo{publisher}{Association for Computing Machinery},
  \bibinfo{address}{New York, NY, USA}, \bibinfo{pages}{5--14}.
\newblock


\bibitem[\protect\citeauthoryear{Attar and Fraenkel}{Attar and
  Fraenkel}{1977}]%
        {Attar:1978}
\bibfield{author}{\bibinfo{person}{R. Attar} {and} \bibinfo{person}{A.~S.
  Fraenkel}.} \bibinfo{year}{1977}\natexlab{}.
\newblock \showarticletitle{Local Feedback in Full-Text Retrieval Systems}.
\newblock \bibinfo{journal}{\emph{J. ACM}} \bibinfo{volume}{24},
  \bibinfo{number}{3} (\bibinfo{date}{jul} \bibinfo{year}{1977}),
  \bibinfo{pages}{397–417}.
\newblock
\showISSN{0004-5411}
\urldef\tempurl%
\url{https://doi.org/10.1145/322017.322021}
\showDOI{\tempurl}


\bibitem[\protect\citeauthoryear{Bahdanau, Cho, and Bengio}{Bahdanau
  et~al\mbox{.}}{2014}]%
        {bahdanau2014neural}
\bibfield{author}{\bibinfo{person}{Dzmitry Bahdanau},
  \bibinfo{person}{Kyunghyun Cho}, {and} \bibinfo{person}{Yoshua Bengio}.}
  \bibinfo{year}{2014}\natexlab{}.
\newblock \showarticletitle{Neural machine translation by jointly learning to
  align and translate}.
\newblock \bibinfo{journal}{\emph{arXiv preprint arXiv:1409.0473}}
  (\bibinfo{year}{2014}).
\newblock


\bibitem[\protect\citeauthoryear{Bender, Gebru, McMillan-Major, and
  Shmitchell}{Bender et~al\mbox{.}}{2021}]%
        {bender:parrots}
\bibfield{author}{\bibinfo{person}{Emily~M. Bender}, \bibinfo{person}{Timnit
  Gebru}, \bibinfo{person}{Angelina McMillan-Major}, {and}
  \bibinfo{person}{Shmargaret Shmitchell}.} \bibinfo{year}{2021}\natexlab{}.
\newblock \showarticletitle{On the Dangers of Stochastic Parrots: Can Language
  Models Be Too Big?}. In \bibinfo{booktitle}{\emph{Proceedings of the 2021 ACM
  Conference on Fairness, Accountability, and Transparency}} (Virtual Event,
  Canada) \emph{(\bibinfo{series}{FAccT '21})}. \bibinfo{publisher}{Association
  for Computing Machinery}, \bibinfo{address}{New York, NY, USA},
  \bibinfo{pages}{610–623}.
\newblock
\showISBNx{9781450383097}
\urldef\tempurl%
\url{https://doi.org/10.1145/3442188.3445922}
\showDOI{\tempurl}


\bibitem[\protect\citeauthoryear{Borgeaud, Mensch, Hoffmann, Cai, Rutherford,
  Millican, van~den Driessche, Lespiau, Damoc, Clark, de~Las~Casas, Guy,
  Menick, Ring, Hennigan, Huang, Maggiore, Jones, Cassirer, Brock, Paganini,
  Irving, Vinyals, Osindero, Simonyan, Rae, Elsen, and Sifre}{Borgeaud
  et~al\mbox{.}}{2021}]%
        {borgeaud:retro}
\bibfield{author}{\bibinfo{person}{Sebastian Borgeaud}, \bibinfo{person}{Arthur
  Mensch}, \bibinfo{person}{Jordan Hoffmann}, \bibinfo{person}{Trevor Cai},
  \bibinfo{person}{Eliza Rutherford}, \bibinfo{person}{Katie Millican},
  \bibinfo{person}{George van~den Driessche}, \bibinfo{person}{Jean-Baptiste
  Lespiau}, \bibinfo{person}{Bogdan Damoc}, \bibinfo{person}{Aidan Clark},
  \bibinfo{person}{Diego de Las~Casas}, \bibinfo{person}{Aurelia Guy},
  \bibinfo{person}{Jacob Menick}, \bibinfo{person}{Roman Ring},
  \bibinfo{person}{Tom Hennigan}, \bibinfo{person}{Saffron Huang},
  \bibinfo{person}{Loren Maggiore}, \bibinfo{person}{Chris Jones},
  \bibinfo{person}{Albin Cassirer}, \bibinfo{person}{Andy Brock},
  \bibinfo{person}{Michela Paganini}, \bibinfo{person}{Geoffrey Irving},
  \bibinfo{person}{Oriol Vinyals}, \bibinfo{person}{Simon Osindero},
  \bibinfo{person}{Karen Simonyan}, \bibinfo{person}{Jack~W. Rae},
  \bibinfo{person}{Erich Elsen}, {and} \bibinfo{person}{Laurent Sifre}.}
  \bibinfo{year}{2021}\natexlab{}.
\newblock \bibinfo{title}{Improving language models by retrieving from
  trillions of tokens}.
\newblock
\newblock
\showeprint[arxiv]{2112.04426}~[cs.CL]


\bibitem[\protect\citeauthoryear{Carlini, Tramer, Wallace, Jagielski,
  Herbert-Voss, Lee, Roberts, Brown, Song, Erlingsson, Oprea, and
  Raffel}{Carlini et~al\mbox{.}}{2021}]%
        {carlini21extracting}
\bibfield{author}{\bibinfo{person}{Nicholas Carlini}, \bibinfo{person}{Florian
  Tramer}, \bibinfo{person}{Eric Wallace}, \bibinfo{person}{Matthew Jagielski},
  \bibinfo{person}{Ariel Herbert-Voss}, \bibinfo{person}{Katherine Lee},
  \bibinfo{person}{Adam Roberts}, \bibinfo{person}{Tom Brown},
  \bibinfo{person}{Dawn Song}, \bibinfo{person}{Ulfar Erlingsson},
  \bibinfo{person}{Alina Oprea}, {and} \bibinfo{person}{Colin Raffel}.}
  \bibinfo{year}{2021}\natexlab{}.
\newblock \showarticletitle{Extracting Training Data from Large Language
  Models}. \bibinfo{howpublished}{arXiv preprint arXiv:2012.07805}. In
  \bibinfo{booktitle}{\emph{USENIX Security Symposium}}.
\newblock
\urldef\tempurl%
\url{https://arxiv.org/abs/2012.07805}
\showURL{%
\tempurl}


\bibitem[\protect\citeauthoryear{Cho, Van~Merri{\"e}nboer, Gulcehre, Bahdanau,
  Bougares, Schwenk, and Bengio}{Cho et~al\mbox{.}}{2014}]%
        {cho2014learning}
\bibfield{author}{\bibinfo{person}{Kyunghyun Cho}, \bibinfo{person}{Bart
  Van~Merri{\"e}nboer}, \bibinfo{person}{Caglar Gulcehre},
  \bibinfo{person}{Dzmitry Bahdanau}, \bibinfo{person}{Fethi Bougares},
  \bibinfo{person}{Holger Schwenk}, {and} \bibinfo{person}{Yoshua Bengio}.}
  \bibinfo{year}{2014}\natexlab{}.
\newblock \showarticletitle{Learning phrase representations using RNN
  encoder-decoder for statistical machine translation}.
\newblock \bibinfo{journal}{\emph{arXiv preprint arXiv:1406.1078}}
  (\bibinfo{year}{2014}).
\newblock


\bibitem[\protect\citeauthoryear{Croft and Harper}{Croft and Harper}{1979}]%
        {Croft:1979}
\bibfield{author}{\bibinfo{person}{W.~B. Croft} {and} \bibinfo{person}{D.~J.
  Harper}.} \bibinfo{year}{1979}\natexlab{}.
\newblock \showarticletitle{{Using Probabilistic Models of Document Retrieval
  Without Relevance Information}}.
\newblock \bibinfo{journal}{\emph{J. of Documentation}} \bibinfo{volume}{35},
  \bibinfo{number}{4} (\bibinfo{year}{1979}), \bibinfo{pages}{285--295}.
\newblock


\bibitem[\protect\citeauthoryear{Dai and Callan}{Dai and Callan}{2019}]%
        {Dai2019}
\bibfield{author}{\bibinfo{person}{Zhuyun Dai} {and} \bibinfo{person}{Jamie
  Callan}.} \bibinfo{year}{2019}\natexlab{}.
\newblock \showarticletitle{Deeper Text Understanding for IR with Contextual
  Neural Language Modeling}. In \bibinfo{booktitle}{\emph{Proceedings of the
  42nd International ACM SIGIR Conference on Research and Development in
  Information Retrieval}} (Paris, France) \emph{(\bibinfo{series}{SIGIR'19})}.
  \bibinfo{publisher}{Association for Computing Machinery},
  \bibinfo{address}{New York, NY, USA}, \bibinfo{pages}{985–988}.
\newblock
\showISBNx{9781450361729}
\urldef\tempurl%
\url{https://doi.org/10.1145/3331184.3331303}
\showDOI{\tempurl}


\bibitem[\protect\citeauthoryear{Das, Godbole, Kavarthapu, Gong, Singhal, Yu,
  Guo, Gao, Zamani, Zaheer, and McCallum}{Das et~al\mbox{.}}{2019}]%
        {das-etal-2019-multi}
\bibfield{author}{\bibinfo{person}{Rajarshi Das}, \bibinfo{person}{Ameya
  Godbole}, \bibinfo{person}{Dilip Kavarthapu}, \bibinfo{person}{Zhiyu Gong},
  \bibinfo{person}{Abhishek Singhal}, \bibinfo{person}{Mo Yu},
  \bibinfo{person}{Xiaoxiao Guo}, \bibinfo{person}{Tian Gao},
  \bibinfo{person}{Hamed Zamani}, \bibinfo{person}{Manzil Zaheer}, {and}
  \bibinfo{person}{Andrew McCallum}.} \bibinfo{year}{2019}\natexlab{}.
\newblock \showarticletitle{Multi-step Entity-centric Information Retrieval for
  Multi-Hop Question Answering}. In \bibinfo{booktitle}{\emph{Proceedings of
  the 2nd Workshop on Machine Reading for Question Answering}}.
  \bibinfo{publisher}{Association for Computational Linguistics},
  \bibinfo{address}{Hong Kong, China}, \bibinfo{pages}{113--118}.
\newblock
\urldef\tempurl%
\url{https://doi.org/10.18653/v1/D19-5816}
\showDOI{\tempurl}


\bibitem[\protect\citeauthoryear{Dehghani, Azarbonyad, Kamps, and
  de~Rijke}{Dehghani et~al\mbox{.}}{2019}]%
        {dehghani2019learning}
\bibfield{author}{\bibinfo{person}{Mostafa Dehghani}, \bibinfo{person}{Hosein
  Azarbonyad}, \bibinfo{person}{Jaap Kamps}, {and} \bibinfo{person}{Maarten de
  Rijke}.} \bibinfo{year}{2019}\natexlab{}.
\newblock \showarticletitle{Learning to transform, combine, and reason in
  open-domain question answering}. In \bibinfo{booktitle}{\emph{Proceedings of
  the Twelfth ACM International Conference on Web Search and Data Mining}}.
  \bibinfo{pages}{681--689}.
\newblock


\bibitem[\protect\citeauthoryear{Dehghani, Zamani, Severyn, Kamps, and
  Croft}{Dehghani et~al\mbox{.}}{2017}]%
        {Dehghani2017Weak}
\bibfield{author}{\bibinfo{person}{Mostafa Dehghani}, \bibinfo{person}{Hamed
  Zamani}, \bibinfo{person}{Aliaksei Severyn}, \bibinfo{person}{Jaap Kamps},
  {and} \bibinfo{person}{W.~Bruce Croft}.} \bibinfo{year}{2017}\natexlab{}.
\newblock \showarticletitle{Neural Ranking Models with Weak Supervision}. In
  \bibinfo{booktitle}{\emph{Proceedings of the 40th International ACM SIGIR
  Conference on Research and Development in Information Retrieval}} (Shinjuku,
  Tokyo, Japan) \emph{(\bibinfo{series}{SIGIR '17})}.
  \bibinfo{publisher}{Association for Computing Machinery},
  \bibinfo{address}{New York, NY, USA}, \bibinfo{pages}{65–74}.
\newblock
\showISBNx{9781450350228}
\urldef\tempurl%
\url{https://doi.org/10.1145/3077136.3080832}
\showDOI{\tempurl}


\bibitem[\protect\citeauthoryear{Fedus, Zoph, and Shazeer}{Fedus
  et~al\mbox{.}}{2021}]%
        {Fedus2021t5xxl}
\bibfield{author}{\bibinfo{person}{William Fedus}, \bibinfo{person}{Barret
  Zoph}, {and} \bibinfo{person}{Noam Shazeer}.}
  \bibinfo{year}{2021}\natexlab{}.
\newblock \showarticletitle{Switch Transformers: Scaling to Trillion Parameter
  Models with Simple and Efficient Sparsity}.
\newblock \bibinfo{journal}{\emph{arXiv:2101.03961}} (\bibinfo{year}{2021}).
\newblock


\bibitem[\protect\citeauthoryear{Feng, Manmatha, and Lavrenko}{Feng
  et~al\mbox{.}}{2004}]%
        {Feng2004}
\bibfield{author}{\bibinfo{person}{S.~L. Feng}, \bibinfo{person}{R. Manmatha},
  {and} \bibinfo{person}{V. Lavrenko}.} \bibinfo{year}{2004}\natexlab{}.
\newblock \showarticletitle{Multiple Bernoulli Relevance Models for Image and
  Video Annotation}. In \bibinfo{booktitle}{\emph{Proceedings of the 2004 IEEE
  Computer Society Conference on Computer Vision and Pattern Recognition}}
  (Washington, D.C., USA) \emph{(\bibinfo{series}{CVPR'04})}.
  \bibinfo{publisher}{IEEE Computer Society}, \bibinfo{address}{USA},
  \bibinfo{pages}{1002–1009}.
\newblock


\bibitem[\protect\citeauthoryear{F{\'e}vry, Baldini~Soares, FitzGerald, Choi,
  and Kwiatkowski}{F{\'e}vry et~al\mbox{.}}{2020}]%
        {fevry-etal-2020-entities}
\bibfield{author}{\bibinfo{person}{Thibault F{\'e}vry}, \bibinfo{person}{Livio
  Baldini~Soares}, \bibinfo{person}{Nicholas FitzGerald},
  \bibinfo{person}{Eunsol Choi}, {and} \bibinfo{person}{Tom Kwiatkowski}.}
  \bibinfo{year}{2020}\natexlab{}.
\newblock \showarticletitle{Entities as Experts: Sparse Memory Access with
  Entity Supervision}. In \bibinfo{booktitle}{\emph{Proceedings of the 2020
  Conference on Empirical Methods in Natural Language Processing (EMNLP)}}.
  \bibinfo{publisher}{Association for Computational Linguistics},
  \bibinfo{address}{Online}, \bibinfo{pages}{4937--4951}.
\newblock
\urldef\tempurl%
\url{https://doi.org/10.18653/v1/2020.emnlp-main.400}
\showDOI{\tempurl}


\bibitem[\protect\citeauthoryear{Fuchsbauer, Ghosal, Hauke, and
  O'Neill}{Fuchsbauer et~al\mbox{.}}{2021}]%
        {Fuchsbauer2021PrivateANN}
\bibfield{author}{\bibinfo{person}{Georg Fuchsbauer}, \bibinfo{person}{Riddhi
  Ghosal}, \bibinfo{person}{Nathan Hauke}, {and} \bibinfo{person}{Adam
  O'Neill}.} \bibinfo{year}{2021}\natexlab{}.
\newblock \bibinfo{title}{Approximate Distance-Comparison-Preserving Symmetric
  Encryption}.
\newblock \bibinfo{howpublished}{Cryptology ePrint Archive, Report 2021/1666}.
\newblock
\newblock
\shownote{\url{https://ia.cr/2021/1666}.}


\bibitem[\protect\citeauthoryear{Goodfellow, Bengio, and Courville}{Goodfellow
  et~al\mbox{.}}{2016}]%
        {Goodfellow-et-al-2016}
\bibfield{author}{\bibinfo{person}{Ian Goodfellow}, \bibinfo{person}{Yoshua
  Bengio}, {and} \bibinfo{person}{Aaron Courville}.}
  \bibinfo{year}{2016}\natexlab{}.
\newblock \bibinfo{booktitle}{\emph{Deep Learning}}.
\newblock \bibinfo{publisher}{MIT Press}.
\newblock
\newblock
\shownote{\url{http://www.deeplearningbook.org}.}


\bibitem[\protect\citeauthoryear{Graves, Wayne, and Danihelka}{Graves
  et~al\mbox{.}}{2014}]%
        {graves2014neural}
\bibfield{author}{\bibinfo{person}{Alex Graves}, \bibinfo{person}{Greg Wayne},
  {and} \bibinfo{person}{Ivo Danihelka}.} \bibinfo{year}{2014}\natexlab{}.
\newblock \showarticletitle{Neural turing machines}.
\newblock \bibinfo{journal}{\emph{arXiv preprint arXiv:1410.5401}}
  (\bibinfo{year}{2014}).
\newblock


\bibitem[\protect\citeauthoryear{Graves, Wayne, Reynolds, Harley, Danihelka,
  Grabska-Barwi{\'n}ska, Colmenarejo, Grefenstette, Ramalho, Agapiou,
  et~al\mbox{.}}{Graves et~al\mbox{.}}{2016}]%
        {graves2016hybrid}
\bibfield{author}{\bibinfo{person}{Alex Graves}, \bibinfo{person}{Greg Wayne},
  \bibinfo{person}{Malcolm Reynolds}, \bibinfo{person}{Tim Harley},
  \bibinfo{person}{Ivo Danihelka}, \bibinfo{person}{Agnieszka
  Grabska-Barwi{\'n}ska}, \bibinfo{person}{Sergio~G{\'o}mez Colmenarejo},
  \bibinfo{person}{Edward Grefenstette}, \bibinfo{person}{Tiago Ramalho},
  \bibinfo{person}{John Agapiou}, {et~al\mbox{.}}}
  \bibinfo{year}{2016}\natexlab{}.
\newblock \showarticletitle{Hybrid computing using a neural network with
  dynamic external memory}.
\newblock \bibinfo{journal}{\emph{Nature}} \bibinfo{volume}{538},
  \bibinfo{number}{7626} (\bibinfo{year}{2016}), \bibinfo{pages}{471--476}.
\newblock


\bibitem[\protect\citeauthoryear{Gur, Neverova, Stauffer, Lim, Kiela, and
  Reiter}{Gur et~al\mbox{.}}{2021}]%
        {Gur+al:2021}
\bibfield{author}{\bibinfo{person}{Shir Gur}, \bibinfo{person}{Natalia
  Neverova}, \bibinfo{person}{Chris Stauffer}, \bibinfo{person}{Ser-Nam Lim},
  \bibinfo{person}{Douwe Kiela}, {and} \bibinfo{person}{Austin Reiter}.}
  \bibinfo{year}{2021}\natexlab{}.
\newblock \showarticletitle{Cross-Modal Retrieval Augmentation for Multi-Modal
  Classification}. In \bibinfo{booktitle}{\emph{Findings of the Association for
  Computational Linguistics: EMNLP 2021}}. \bibinfo{publisher}{Association for
  Computational Linguistics}, \bibinfo{address}{Punta Cana, Dominican
  Republic}, \bibinfo{pages}{111--123}.
\newblock
\urldef\tempurl%
\url{https://doi.org/10.18653/v1/2021.findings-emnlp.11}
\showDOI{\tempurl}


\bibitem[\protect\citeauthoryear{Guu, Lee, Tung, Pasupat, and Chang}{Guu
  et~al\mbox{.}}{2020}]%
        {Guu+al:2020}
\bibfield{author}{\bibinfo{person}{Kelvin Guu}, \bibinfo{person}{Kenton Lee},
  \bibinfo{person}{Zora Tung}, \bibinfo{person}{Panupong Pasupat}, {and}
  \bibinfo{person}{Ming{-}Wei Chang}.} \bibinfo{year}{2020}\natexlab{}.
\newblock \showarticletitle{Retrieval Augmented Language Model Pre-Training}.
  In \bibinfo{booktitle}{\emph{Proceedings of the 37th International Conference
  on Machine Learning, {ICML} 2020, 13-18 July 2020, Virtual Event}}
  \emph{(\bibinfo{series}{Proceedings of Machine Learning Research},
  Vol.~\bibinfo{volume}{119})}. \bibinfo{publisher}{{PMLR}},
  \bibinfo{pages}{3929--3938}.
\newblock
\urldef\tempurl%
\url{http://proceedings.mlr.press/v119/guu20a.html}
\showURL{%
\tempurl}


\bibitem[\protect\citeauthoryear{Hashemi, Zamani, and Croft}{Hashemi
  et~al\mbox{.}}{2020}]%
        {Hashemi2020GT}
\bibfield{author}{\bibinfo{person}{Helia Hashemi}, \bibinfo{person}{Hamed
  Zamani}, {and} \bibinfo{person}{W.~Bruce Croft}.}
  \bibinfo{year}{2020}\natexlab{}.
\newblock \showarticletitle{Guided Transformer: Leveraging Multiple External
  Sources for Representation Learning in Conversational Search}. In
  \bibinfo{booktitle}{\emph{Proceedings of the 43rd International ACM SIGIR
  Conference on Research and Development in Information Retrieval}}
  \emph{(\bibinfo{series}{SIGIR '20})}. \bibinfo{publisher}{Association for
  Computing Machinery}, \bibinfo{address}{New York, NY, USA},
  \bibinfo{pages}{1131–1140}.
\newblock
\showISBNx{9781450380164}
\urldef\tempurl%
\url{https://doi.org/10.1145/3397271.3401061}
\showURL{%
\tempurl}


\bibitem[\protect\citeauthoryear{Hashemi, Zamani, and Croft}{Hashemi
  et~al\mbox{.}}{2021}]%
        {Hashemi2021NMIR}
\bibfield{author}{\bibinfo{person}{Helia Hashemi}, \bibinfo{person}{Hamed
  Zamani}, {and} \bibinfo{person}{W.~Bruce Croft}.}
  \bibinfo{year}{2021}\natexlab{}.
\newblock \bibinfo{booktitle}{\emph{Learning Multiple Intent Representations
  for Search Queries}}.
\newblock \bibinfo{publisher}{Association for Computing Machinery},
  \bibinfo{address}{New York, NY, USA}, \bibinfo{pages}{669–679}.
\newblock
\showISBNx{9781450384469}
\urldef\tempurl%
\url{https://doi.org/10.1145/3459637.3482445}
\showURL{%
\tempurl}


\bibitem[\protect\citeauthoryear{Hashimoto, Guu, Oren, and Liang}{Hashimoto
  et~al\mbox{.}}{2018}]%
        {Hashimoto:2018}
\bibfield{author}{\bibinfo{person}{Tatsunori~B Hashimoto},
  \bibinfo{person}{Kelvin Guu}, \bibinfo{person}{Yonatan Oren}, {and}
  \bibinfo{person}{Percy~S Liang}.} \bibinfo{year}{2018}\natexlab{}.
\newblock \showarticletitle{A Retrieve-and-Edit Framework for Predicting
  Structured Outputs}. In \bibinfo{booktitle}{\emph{Advances in Neural
  Information Processing Systems}},
  \bibfield{editor}{\bibinfo{person}{S.~Bengio}, \bibinfo{person}{H.~Wallach},
  \bibinfo{person}{H.~Larochelle}, \bibinfo{person}{K.~Grauman},
  \bibinfo{person}{N.~Cesa-Bianchi}, {and} \bibinfo{person}{R.~Garnett}}
  (Eds.), Vol.~\bibinfo{volume}{31}. \bibinfo{publisher}{Curran Associates,
  Inc.}
\newblock
\urldef\tempurl%
\url{https://proceedings.neurips.cc/paper/2018/file/cd17d3ce3b64f227987cd92cd701cc58-Paper.pdf}
\showURL{%
\tempurl}


\bibitem[\protect\citeauthoryear{Hochreiter and Schmidhuber}{Hochreiter and
  Schmidhuber}{1997}]%
        {hochreiter1997long}
\bibfield{author}{\bibinfo{person}{Sepp Hochreiter} {and}
  \bibinfo{person}{J{\"u}rgen Schmidhuber}.} \bibinfo{year}{1997}\natexlab{}.
\newblock \showarticletitle{Long short-term memory}.
\newblock \bibinfo{journal}{\emph{Neural computation}} \bibinfo{volume}{9},
  \bibinfo{number}{8} (\bibinfo{year}{1997}), \bibinfo{pages}{1735--1780}.
\newblock


\bibitem[\protect\citeauthoryear{Hofst\"{a}tter, Mitra, Zamani, Craswell, and
  Hanbury}{Hofst\"{a}tter et~al\mbox{.}}{2021}]%
        {Hofstatter2021ICDM}
\bibfield{author}{\bibinfo{person}{Sebastian Hofst\"{a}tter},
  \bibinfo{person}{Bhaskar Mitra}, \bibinfo{person}{Hamed Zamani},
  \bibinfo{person}{Nick Craswell}, {and} \bibinfo{person}{Allan Hanbury}.}
  \bibinfo{year}{2021}\natexlab{}.
\newblock \bibinfo{booktitle}{\emph{Intra-Document Cascading: Learning to
  Select Passages for Neural Document Ranking}}.
\newblock \bibinfo{publisher}{Association for Computing Machinery},
  \bibinfo{address}{New York, NY, USA}, \bibinfo{pages}{1349–1358}.
\newblock
\showISBNx{9781450380379}
\urldef\tempurl%
\url{https://doi.org/10.1145/3404835.3462889}
\showURL{%
\tempurl}


\bibitem[\protect\citeauthoryear{Izacard and Grave}{Izacard and Grave}{2020}]%
        {izacard2020leveraging}
\bibfield{author}{\bibinfo{person}{Gautier Izacard} {and}
  \bibinfo{person}{Edouard Grave}.} \bibinfo{year}{2020}\natexlab{}.
\newblock \showarticletitle{Leveraging passage retrieval with generative models
  for open domain question answering}.
\newblock \bibinfo{journal}{\emph{arXiv preprint arXiv:2007.01282}}
  (\bibinfo{year}{2020}).
\newblock


\bibitem[\protect\citeauthoryear{Jeon, Lavrenko, and Manmatha}{Jeon
  et~al\mbox{.}}{2003}]%
        {jeonSIGIR03}
\bibfield{author}{\bibinfo{person}{J. Jeon}, \bibinfo{person}{V. Lavrenko},
  {and} \bibinfo{person}{R. Manmatha}.} \bibinfo{year}{2003}\natexlab{}.
\newblock \showarticletitle{Automatic Image Annotation and Retrieval Using
  Cross-Media Relevance Models} \emph{(\bibinfo{series}{SIGIR '03})}.
  \bibinfo{publisher}{Association for Computing Machinery},
  \bibinfo{address}{New York, NY, USA}, \bibinfo{pages}{119–126}.
\newblock
\showISBNx{1581136463}
\urldef\tempurl%
\url{https://doi.org/10.1145/860435.860459}
\showDOI{\tempurl}


\bibitem[\protect\citeauthoryear{Kassner and Sch{\"u}tze}{Kassner and
  Sch{\"u}tze}{2020}]%
        {kassner-schutze-2020-bert}
\bibfield{author}{\bibinfo{person}{Nora Kassner} {and} \bibinfo{person}{Hinrich
  Sch{\"u}tze}.} \bibinfo{year}{2020}\natexlab{}.
\newblock \showarticletitle{{BERT}-k{NN}: Adding a k{NN} Search Component to
  Pretrained Language Models for Better {QA}}. In
  \bibinfo{booktitle}{\emph{Findings of the Association for Computational
  Linguistics: EMNLP 2020}}. \bibinfo{publisher}{Association for Computational
  Linguistics}, \bibinfo{address}{Online}, \bibinfo{pages}{3424--3430}.
\newblock
\urldef\tempurl%
\url{https://doi.org/10.18653/v1/2020.findings-emnlp.307}
\showDOI{\tempurl}


\bibitem[\protect\citeauthoryear{Kelly, Arguello, and Capra}{Kelly
  et~al\mbox{.}}{2013}]%
        {kelly:task-workshop}
\bibfield{author}{\bibinfo{person}{Diane Kelly}, \bibinfo{person}{Jaime
  Arguello}, {and} \bibinfo{person}{Robert Capra}.}
  \bibinfo{year}{2013}\natexlab{}.
\newblock \showarticletitle{NSF Workshop on Task-based Information Search
  Systems}.
\newblock \bibinfo{journal}{\emph{SIGIR Forum}} \bibinfo{volume}{47},
  \bibinfo{number}{2} (\bibinfo{year}{2013}).
\newblock


\bibitem[\protect\citeauthoryear{Khandelwal, Fan, Jurafsky, Zettlemoyer, and
  Lewis}{Khandelwal et~al\mbox{.}}{2021}]%
        {khandelwal2021nearest}
\bibfield{author}{\bibinfo{person}{Urvashi Khandelwal}, \bibinfo{person}{Angela
  Fan}, \bibinfo{person}{Dan Jurafsky}, \bibinfo{person}{Luke Zettlemoyer},
  {and} \bibinfo{person}{Mike Lewis}.} \bibinfo{year}{2021}\natexlab{}.
\newblock \showarticletitle{Nearest Neighbor Machine Translation}. In
  \bibinfo{booktitle}{\emph{International Conference on Learning
  Representations}}.
\newblock
\urldef\tempurl%
\url{https://openreview.net/forum?id=7wCBOfJ8hJM}
\showURL{%
\tempurl}


\bibitem[\protect\citeauthoryear{Khandelwal, Levy, Jurafsky, Zettlemoyer, and
  Lewis}{Khandelwal et~al\mbox{.}}{2020}]%
        {khandelwal:knnlm}
\bibfield{author}{\bibinfo{person}{Urvashi Khandelwal}, \bibinfo{person}{Omer
  Levy}, \bibinfo{person}{Dan Jurafsky}, \bibinfo{person}{Luke Zettlemoyer},
  {and} \bibinfo{person}{Mike Lewis}.} \bibinfo{year}{2020}\natexlab{}.
\newblock \showarticletitle{Generalization through Memorization: Nearest
  Neighbor Language Models}. In \bibinfo{booktitle}{\emph{International
  Conference on Learning Representations}}.
\newblock
\urldef\tempurl%
\url{https://openreview.net/forum?id=HklBjCEKvH}
\showURL{%
\tempurl}


\bibitem[\protect\citeauthoryear{Khattab, Potts, and Zaharia}{Khattab
  et~al\mbox{.}}{2020}]%
        {khattabNEURIPS21}
\bibfield{author}{\bibinfo{person}{Omar Khattab}, \bibinfo{person}{Christopher
  Potts}, {and} \bibinfo{person}{Matei Zaharia}.}
  \bibinfo{year}{2020}\natexlab{}.
\newblock \showarticletitle{Baleen: Robust Multi-Hop Reasoning at Scale via
  Condensed Retrieval}. In \bibinfo{booktitle}{\emph{Advances in Neural
  Information Processing Systems 33: Annual Conference on Neural Information
  Processing Systems 2020, NeurIPS 2020, December 6-12, 2020, virtual}}.
\newblock


\bibitem[\protect\citeauthoryear{Komeili, Shuster, and Weston}{Komeili
  et~al\mbox{.}}{2021}]%
        {komeili:blenderbot2}
\bibfield{author}{\bibinfo{person}{Mojtaba Komeili}, \bibinfo{person}{Kurt
  Shuster}, {and} \bibinfo{person}{Jason Weston}.}
  \bibinfo{year}{2021}\natexlab{}.
\newblock \showarticletitle{Internet-Augmented Dialogue Generation}.
\newblock \bibinfo{journal}{\emph{CoRR}}  \bibinfo{volume}{abs/2107.07566}
  (\bibinfo{year}{2021}).
\newblock
\showeprint[arXiv]{2107.07566}
\urldef\tempurl%
\url{https://arxiv.org/abs/2107.07566}
\showURL{%
\tempurl}


\bibitem[\protect\citeauthoryear{Kuo, Angelova, Lin, and Dai}{Kuo
  et~al\mbox{.}}{2020}]%
        {kuo2020mask2cad}
\bibfield{author}{\bibinfo{person}{Weicheng Kuo}, \bibinfo{person}{Anelia
  Angelova}, \bibinfo{person}{Tsung-Yi Lin}, {and} \bibinfo{person}{Angela
  Dai}.} \bibinfo{year}{2020}\natexlab{}.
\newblock \showarticletitle{Mask2cad: 3d shape prediction by learning to
  segment and retrieve}. In \bibinfo{booktitle}{\emph{European Conference on
  Computer Vision}}. Springer, \bibinfo{pages}{260--277}.
\newblock


\bibitem[\protect\citeauthoryear{Lavrenko and Croft}{Lavrenko and
  Croft}{2001}]%
        {Lavrenko2001RM}
\bibfield{author}{\bibinfo{person}{Victor Lavrenko} {and}
  \bibinfo{person}{W.~Bruce Croft}.} \bibinfo{year}{2001}\natexlab{}.
\newblock \showarticletitle{Relevance Based Language Models}. In
  \bibinfo{booktitle}{\emph{Proceedings of the 24th Annual International ACM
  SIGIR Conference on Research and Development in Information Retrieval}} (New
  Orleans, Louisiana, USA) \emph{(\bibinfo{series}{SIGIR '01})}.
  \bibinfo{publisher}{Association for Computing Machinery},
  \bibinfo{address}{New York, NY, USA}, \bibinfo{pages}{120–127}.
\newblock
\showISBNx{1581133316}
\urldef\tempurl%
\url{https://doi.org/10.1145/383952.383972}
\showDOI{\tempurl}


\bibitem[\protect\citeauthoryear{Lee, Chang, and Toutanova}{Lee
  et~al\mbox{.}}{2019}]%
        {lee2019latent}
\bibfield{author}{\bibinfo{person}{Kenton Lee}, \bibinfo{person}{Ming-Wei
  Chang}, {and} \bibinfo{person}{Kristina Toutanova}.}
  \bibinfo{year}{2019}\natexlab{}.
\newblock \showarticletitle{Latent retrieval for weakly supervised open domain
  question answering}.
\newblock \bibinfo{journal}{\emph{arXiv preprint arXiv:1906.00300}}
  (\bibinfo{year}{2019}).
\newblock


\bibitem[\protect\citeauthoryear{Lee, Sung, Yu, and Kim}{Lee
  et~al\mbox{.}}{2018}]%
        {lee2018memory}
\bibfield{author}{\bibinfo{person}{Sangho Lee}, \bibinfo{person}{Jinyoung
  Sung}, \bibinfo{person}{Youngjae Yu}, {and} \bibinfo{person}{Gunhee Kim}.}
  \bibinfo{year}{2018}\natexlab{}.
\newblock \showarticletitle{A memory network approach for story-based temporal
  summarization of 360 videos}. In \bibinfo{booktitle}{\emph{Proceedings of the
  IEEE Conference on Computer Vision and Pattern Recognition}}.
  \bibinfo{pages}{1410--1419}.
\newblock


\bibitem[\protect\citeauthoryear{Lewis, Perez, Piktus, Petroni, Karpukhin,
  Goyal, K{\"{u}}ttler, Lewis, Yih, Rockt{\"{a}}schel, Riedel, and Kiela}{Lewis
  et~al\mbox{.}}{2020}]%
        {Lewis+al:2020}
\bibfield{author}{\bibinfo{person}{Patrick S.~H. Lewis}, \bibinfo{person}{Ethan
  Perez}, \bibinfo{person}{Aleksandra Piktus}, \bibinfo{person}{Fabio Petroni},
  \bibinfo{person}{Vladimir Karpukhin}, \bibinfo{person}{Naman Goyal},
  \bibinfo{person}{Heinrich K{\"{u}}ttler}, \bibinfo{person}{Mike Lewis},
  \bibinfo{person}{Wen{-}tau Yih}, \bibinfo{person}{Tim Rockt{\"{a}}schel},
  \bibinfo{person}{Sebastian Riedel}, {and} \bibinfo{person}{Douwe Kiela}.}
  \bibinfo{year}{2020}\natexlab{}.
\newblock \showarticletitle{Retrieval-Augmented Generation for
  Knowledge-Intensive {NLP} Tasks}. In \bibinfo{booktitle}{\emph{Advances in
  Neural Information Processing Systems 33: Annual Conference on Neural
  Information Processing Systems 2020, NeurIPS 2020, December 6-12, 2020,
  virtual}}, \bibfield{editor}{\bibinfo{person}{Hugo Larochelle},
  \bibinfo{person}{Marc'Aurelio Ranzato}, \bibinfo{person}{Raia Hadsell},
  \bibinfo{person}{Maria{-}Florina Balcan}, {and} \bibinfo{person}{Hsuan{-}Tien
  Lin}} (Eds.).
\newblock
\urldef\tempurl%
\url{https://proceedings.neurips.cc/paper/2020/hash/6b493230205f780e1bc26945df7481e5-Abstract.html}
\showURL{%
\tempurl}


\bibitem[\protect\citeauthoryear{Li, Dai, Guibas, and Nie{\ss}ner}{Li
  et~al\mbox{.}}{2015}]%
        {li2015database}
\bibfield{author}{\bibinfo{person}{Yangyan Li}, \bibinfo{person}{Angela Dai},
  \bibinfo{person}{Leonidas Guibas}, {and} \bibinfo{person}{Matthias
  Nie{\ss}ner}.} \bibinfo{year}{2015}\natexlab{}.
\newblock \showarticletitle{Database-assisted object retrieval for real-time 3d
  reconstruction}. In \bibinfo{booktitle}{\emph{Computer graphics forum}},
  Vol.~\bibinfo{volume}{34}. Wiley Online Library, \bibinfo{pages}{435--446}.
\newblock


\bibitem[\protect\citeauthoryear{Li, Liu, Xiong, and Liu}{Li
  et~al\mbox{.}}{2021}]%
        {Li2021DANCE}
\bibfield{author}{\bibinfo{person}{Yizhi Li}, \bibinfo{person}{Zhenghao Liu},
  \bibinfo{person}{Chenyan Xiong}, {and} \bibinfo{person}{Zhiyuan Liu}.}
  \bibinfo{year}{2021}\natexlab{}.
\newblock \bibinfo{booktitle}{\emph{More Robust Dense Retrieval with
  Contrastive Dual Learning}}.
\newblock \bibinfo{publisher}{Association for Computing Machinery},
  \bibinfo{address}{New York, NY, USA}, \bibinfo{pages}{287–296}.
\newblock
\showISBNx{9781450386111}
\urldef\tempurl%
\url{https://doi.org/10.1145/3471158.3472245}
\showURL{%
\tempurl}


\bibitem[\protect\citeauthoryear{Liu}{Liu}{2009}]%
        {Liu2009LTR}
\bibfield{author}{\bibinfo{person}{Tie-Yan Liu}.}
  \bibinfo{year}{2009}\natexlab{}.
\newblock \showarticletitle{Learning to Rank for Information Retrieval}.
\newblock \bibinfo{journal}{\emph{Found. Trends Inf. Retr.}}
  \bibinfo{volume}{3}, \bibinfo{number}{3} (\bibinfo{date}{mar}
  \bibinfo{year}{2009}), \bibinfo{pages}{225–331}.
\newblock
\showISSN{1554-0669}
\urldef\tempurl%
\url{https://doi.org/10.1561/1500000016}
\showDOI{\tempurl}


\bibitem[\protect\citeauthoryear{Oh, Lee, Xu, and Kim}{Oh
  et~al\mbox{.}}{2019}]%
        {oh2019video}
\bibfield{author}{\bibinfo{person}{Seoung~Wug Oh}, \bibinfo{person}{Joon-Young
  Lee}, \bibinfo{person}{Ning Xu}, {and} \bibinfo{person}{Seon~Joo Kim}.}
  \bibinfo{year}{2019}\natexlab{}.
\newblock \showarticletitle{Video object segmentation using space-time memory
  networks}. In \bibinfo{booktitle}{\emph{Proceedings of the IEEE/CVF
  International Conference on Computer Vision}}. \bibinfo{pages}{9226--9235}.
\newblock


\bibitem[\protect\citeauthoryear{Pasupat, Zhang, and Guu}{Pasupat
  et~al\mbox{.}}{2021}]%
        {Pasupat+al:2021}
\bibfield{author}{\bibinfo{person}{Panupong Pasupat}, \bibinfo{person}{Yuan
  Zhang}, {and} \bibinfo{person}{Kelvin Guu}.} \bibinfo{year}{2021}\natexlab{}.
\newblock \showarticletitle{Controllable Semantic Parsing via Retrieval
  Augmentation}. In \bibinfo{booktitle}{\emph{Proceedings of the 2021
  Conference on Empirical Methods in Natural Language Processing, {EMNLP} 2021,
  Virtual Event / Punta Cana, Dominican Republic, 7-11 November, 2021}},
  \bibfield{editor}{\bibinfo{person}{Marie{-}Francine Moens},
  \bibinfo{person}{Xuanjing Huang}, \bibinfo{person}{Lucia Specia}, {and}
  \bibinfo{person}{Scott~Wen{-}tau Yih}} (Eds.).
  \bibinfo{publisher}{Association for Computational Linguistics},
  \bibinfo{pages}{7683--7698}.
\newblock
\urldef\tempurl%
\url{https://aclanthology.org/2021.emnlp-main.607}
\showURL{%
\tempurl}


\bibitem[\protect\citeauthoryear{Peters, Neumann, Iyyer, Gardner, Clark, Lee,
  and Zettlemoyer}{Peters et~al\mbox{.}}{2018}]%
        {peters2018elmo}
\bibfield{author}{\bibinfo{person}{Matthew~E. Peters}, \bibinfo{person}{Mark
  Neumann}, \bibinfo{person}{Mohit Iyyer}, \bibinfo{person}{Matt Gardner},
  \bibinfo{person}{Christopher Clark}, \bibinfo{person}{Kenton Lee}, {and}
  \bibinfo{person}{Luke Zettlemoyer}.} \bibinfo{year}{2018}\natexlab{}.
\newblock \showarticletitle{Deep Contextualized Word Representations}. In
  \bibinfo{booktitle}{\emph{Proceedings of the 2018 Conference of the North
  {A}merican Chapter of the Association for Computational Linguistics: Human
  Language Technologies, Volume 1 (Long Papers)}}.
  \bibinfo{publisher}{Association for Computational Linguistics},
  \bibinfo{address}{New Orleans, Louisiana}, \bibinfo{pages}{2227--2237}.
\newblock
\urldef\tempurl%
\url{https://doi.org/10.18653/v1/N18-1202}
\showDOI{\tempurl}


\bibitem[\protect\citeauthoryear{Prakash, Killingback, and Zamani}{Prakash
  et~al\mbox{.}}{2021}]%
        {Prakash2021RANCE}
\bibfield{author}{\bibinfo{person}{Prafull Prakash}, \bibinfo{person}{Julian
  Killingback}, {and} \bibinfo{person}{Hamed Zamani}.}
  \bibinfo{year}{2021}\natexlab{}.
\newblock \bibinfo{booktitle}{\emph{Learning Robust Dense Retrieval Models from
  Incomplete Relevance Labels}}.
\newblock \bibinfo{publisher}{Association for Computing Machinery},
  \bibinfo{address}{New York, NY, USA}, \bibinfo{pages}{1728–1732}.
\newblock
\showISBNx{9781450380379}
\urldef\tempurl%
\url{https://doi.org/10.1145/3404835.3463106}
\showURL{%
\tempurl}


\bibitem[\protect\citeauthoryear{Qu, Zamani, Yang, Croft, and
  Learned-Miller}{Qu et~al\mbox{.}}{2021}]%
        {Qu2021OKVQA}
\bibfield{author}{\bibinfo{person}{Chen Qu}, \bibinfo{person}{Hamed Zamani},
  \bibinfo{person}{Liu Yang}, \bibinfo{person}{W.~Bruce Croft}, {and}
  \bibinfo{person}{Erik Learned-Miller}.} \bibinfo{year}{2021}\natexlab{}.
\newblock \bibinfo{booktitle}{\emph{Passage Retrieval for Outside-Knowledge
  Visual Question Answering}}.
\newblock \bibinfo{publisher}{Association for Computing Machinery},
  \bibinfo{address}{New York, NY, USA}, \bibinfo{pages}{1753–1757}.
\newblock
\showISBNx{9781450380379}
\urldef\tempurl%
\url{https://doi.org/10.1145/3404835.3462987}
\showURL{%
\tempurl}


\bibitem[\protect\citeauthoryear{Radford, Kim, Hallacy, Ramesh, Goh, Agarwal,
  Sastry, Askell, Mishkin, Clark, et~al\mbox{.}}{Radford et~al\mbox{.}}{2021}]%
        {radford2021learning}
\bibfield{author}{\bibinfo{person}{Alec Radford}, \bibinfo{person}{Jong~Wook
  Kim}, \bibinfo{person}{Chris Hallacy}, \bibinfo{person}{Aditya Ramesh},
  \bibinfo{person}{Gabriel Goh}, \bibinfo{person}{Sandhini Agarwal},
  \bibinfo{person}{Girish Sastry}, \bibinfo{person}{Amanda Askell},
  \bibinfo{person}{Pamela Mishkin}, \bibinfo{person}{Jack Clark},
  {et~al\mbox{.}}} \bibinfo{year}{2021}\natexlab{}.
\newblock \showarticletitle{Learning transferable visual models from natural
  language supervision}. In \bibinfo{booktitle}{\emph{International Conference
  on Machine Learning}}. PMLR, \bibinfo{pages}{8748--8763}.
\newblock


\bibitem[\protect\citeauthoryear{Robertson}{Robertson}{1997}]%
        {Robertson1997PRP}
\bibfield{author}{\bibinfo{person}{Stephen~E. Robertson}.}
  \bibinfo{year}{1997}\natexlab{}.
\newblock \bibinfo{booktitle}{\emph{The Probability Ranking Principle in IR}}.
\newblock \bibinfo{publisher}{Morgan Kaufmann Publishers Inc.},
  \bibinfo{address}{San Francisco, CA, USA}, \bibinfo{pages}{281–286}.
\newblock
\showISBNx{1558604545}


\bibitem[\protect\citeauthoryear{Roller, Dinan, Goyal, Ju, Williamson, Liu, Xu,
  Ott, Smith, Boureau, and Weston}{Roller et~al\mbox{.}}{2021}]%
        {roller:blenderbot}
\bibfield{author}{\bibinfo{person}{Stephen Roller}, \bibinfo{person}{Emily
  Dinan}, \bibinfo{person}{Naman Goyal}, \bibinfo{person}{Da Ju},
  \bibinfo{person}{Mary Williamson}, \bibinfo{person}{Yinhan Liu},
  \bibinfo{person}{Jing Xu}, \bibinfo{person}{Myle Ott},
  \bibinfo{person}{Eric~Michael Smith}, \bibinfo{person}{Y-Lan Boureau}, {and}
  \bibinfo{person}{Jason Weston}.} \bibinfo{year}{2021}\natexlab{}.
\newblock \showarticletitle{Recipes for Building an Open-Domain Chatbot}. In
  \bibinfo{booktitle}{\emph{Proceedings of the 16th Conference of the European
  Chapter of the Association for Computational Linguistics: Main Volume}}.
  \bibinfo{publisher}{Association for Computational Linguistics},
  \bibinfo{address}{Online}, \bibinfo{pages}{300--325}.
\newblock
\urldef\tempurl%
\url{https://doi.org/10.18653/v1/2021.eacl-main.24}
\showDOI{\tempurl}


\bibitem[\protect\citeauthoryear{Santoro, Bartunov, Botvinick, Wierstra, and
  Lillicrap}{Santoro et~al\mbox{.}}{2016}]%
        {santoro2016meta}
\bibfield{author}{\bibinfo{person}{Adam Santoro}, \bibinfo{person}{Sergey
  Bartunov}, \bibinfo{person}{Matthew Botvinick}, \bibinfo{person}{Daan
  Wierstra}, {and} \bibinfo{person}{Timothy Lillicrap}.}
  \bibinfo{year}{2016}\natexlab{}.
\newblock \showarticletitle{Meta-learning with memory-augmented neural
  networks}. In \bibinfo{booktitle}{\emph{International conference on machine
  learning}}. PMLR, \bibinfo{pages}{1842--1850}.
\newblock


\bibitem[\protect\citeauthoryear{Saracevic}{Saracevic}{1996}]%
        {Saracevic1996relevance}
\bibfield{author}{\bibinfo{person}{Tefko Saracevic}.}
  \bibinfo{year}{1996}\natexlab{}.
\newblock \showarticletitle{Relevance reconsidered}. In
  \bibinfo{booktitle}{\emph{Proceedings of the Second Conference on Conceptions
  of Library and Information Science}} (Copenhagen, Denmark).
\newblock


\bibitem[\protect\citeauthoryear{Shuster, Poff, Chen, Kiela, and
  Weston}{Shuster et~al\mbox{.}}{2021}]%
        {shuster2021retrieval}
\bibfield{author}{\bibinfo{person}{Kurt Shuster}, \bibinfo{person}{Spencer
  Poff}, \bibinfo{person}{Moya Chen}, \bibinfo{person}{Douwe Kiela}, {and}
  \bibinfo{person}{Jason Weston}.} \bibinfo{year}{2021}\natexlab{}.
\newblock \showarticletitle{Retrieval augmentation reduces hallucination in
  conversation}.
\newblock \bibinfo{journal}{\emph{arXiv preprint arXiv:2104.07567}}
  (\bibinfo{year}{2021}).
\newblock


\bibitem[\protect\citeauthoryear{Siddiqui, Thies, Ma, Shan, Nießner, and
  Dai}{Siddiqui et~al\mbox{.}}{2021}]%
        {siddiqui2021retrievalfuse}
\bibfield{author}{\bibinfo{person}{Yawar Siddiqui}, \bibinfo{person}{Justus
  Thies}, \bibinfo{person}{Fangchang Ma}, \bibinfo{person}{Qi Shan},
  \bibinfo{person}{Matthias Nießner}, {and} \bibinfo{person}{Angela Dai}.}
  \bibinfo{year}{2021}\natexlab{}.
\newblock \bibinfo{title}{RetrievalFuse: Neural 3D Scene Reconstruction with a
  Database}.
\newblock
\newblock
\showeprint[arxiv]{2104.00024}~[cs.CV]


\bibitem[\protect\citeauthoryear{Sukhbaatar, Weston, Fergus,
  et~al\mbox{.}}{Sukhbaatar et~al\mbox{.}}{2015}]%
        {sukhbaatar2015end}
\bibfield{author}{\bibinfo{person}{Sainbayar Sukhbaatar},
  \bibinfo{person}{Jason Weston}, \bibinfo{person}{Rob Fergus},
  {et~al\mbox{.}}} \bibinfo{year}{2015}\natexlab{}.
\newblock \showarticletitle{End-to-end memory networks}.
\newblock \bibinfo{journal}{\emph{Advances in neural information processing
  systems}}  \bibinfo{volume}{28} (\bibinfo{year}{2015}).
\newblock


\bibitem[\protect\citeauthoryear{Sutskever, Vinyals, and Le}{Sutskever
  et~al\mbox{.}}{2014}]%
        {Sutskever:2014:seq2seq}
\bibfield{author}{\bibinfo{person}{Ilya Sutskever}, \bibinfo{person}{Oriol
  Vinyals}, {and} \bibinfo{person}{Quoc~V Le}.}
  \bibinfo{year}{2014}\natexlab{}.
\newblock \showarticletitle{Sequence to Sequence Learning with Neural
  Networks}. In \bibinfo{booktitle}{\emph{Advances in Neural Information
  Processing Systems}}, \bibfield{editor}{\bibinfo{person}{Z.~Ghahramani},
  \bibinfo{person}{M.~Welling}, \bibinfo{person}{C.~Cortes},
  \bibinfo{person}{N.~Lawrence}, {and} \bibinfo{person}{K.~Q. Weinberger}}
  (Eds.), Vol.~\bibinfo{volume}{27}. \bibinfo{publisher}{Curran Associates,
  Inc.}
\newblock


\bibitem[\protect\citeauthoryear{Tay, Tran, Dehghani, Ni, Bahri, Mehta, Qin,
  Hui, Zhao, Gupta, et~al\mbox{.}}{Tay et~al\mbox{.}}{2022}]%
        {tay2022transformer}
\bibfield{author}{\bibinfo{person}{Yi Tay}, \bibinfo{person}{Vinh~Q Tran},
  \bibinfo{person}{Mostafa Dehghani}, \bibinfo{person}{Jianmo Ni},
  \bibinfo{person}{Dara Bahri}, \bibinfo{person}{Harsh Mehta},
  \bibinfo{person}{Zhen Qin}, \bibinfo{person}{Kai Hui}, \bibinfo{person}{Zhe
  Zhao}, \bibinfo{person}{Jai Gupta}, {et~al\mbox{.}}}
  \bibinfo{year}{2022}\natexlab{}.
\newblock \showarticletitle{Transformer memory as a differentiable search
  index}.
\newblock \bibinfo{journal}{\emph{arXiv preprint arXiv:2202.06991}}
  (\bibinfo{year}{2022}).
\newblock


\bibitem[\protect\citeauthoryear{Thoppilan, Freitas, Hall, Shazeer,
  Kulshreshtha, Cheng, Jin, Bos, Baker, Du, Li, Lee, Zheng, Ghafouri, Menegali,
  Huang, Krikun, Lepikhin, Qin, Chen, Xu, Chen, Roberts, Bosma, Zhou, Chang,
  Krivokon, Rusch, Pickett, Meier-Hellstern, Morris, Doshi, Santos, Duke,
  Soraker, Zevenbergen, Prabhakaran, Diaz, Hutchinson, Olson, Molina,
  Hoffman-John, Lee, Aroyo, Rajakumar, Butryna, Lamm, Kuzmina, Fenton, Cohen,
  Bernstein, Kurzweil, Aguera-Arcas, Cui, Croak, Chi, and Le}{Thoppilan
  et~al\mbox{.}}{2022}]%
        {Thoppilan:2022}
\bibfield{author}{\bibinfo{person}{Romal Thoppilan}, \bibinfo{person}{Daniel~De
  Freitas}, \bibinfo{person}{Jamie Hall}, \bibinfo{person}{Noam Shazeer},
  \bibinfo{person}{Apoorv Kulshreshtha}, \bibinfo{person}{Heng-Tze Cheng},
  \bibinfo{person}{Alicia Jin}, \bibinfo{person}{Taylor Bos},
  \bibinfo{person}{Leslie Baker}, \bibinfo{person}{Yu Du},
  \bibinfo{person}{YaGuang Li}, \bibinfo{person}{Hongrae Lee},
  \bibinfo{person}{Huaixiu~Steven Zheng}, \bibinfo{person}{Amin Ghafouri},
  \bibinfo{person}{Marcelo Menegali}, \bibinfo{person}{Yanping Huang},
  \bibinfo{person}{Maxim Krikun}, \bibinfo{person}{Dmitry Lepikhin},
  \bibinfo{person}{James Qin}, \bibinfo{person}{Dehao Chen},
  \bibinfo{person}{Yuanzhong Xu}, \bibinfo{person}{Zhifeng Chen},
  \bibinfo{person}{Adam Roberts}, \bibinfo{person}{Maarten Bosma},
  \bibinfo{person}{Yanqi Zhou}, \bibinfo{person}{Chung-Ching Chang},
  \bibinfo{person}{Igor Krivokon}, \bibinfo{person}{Will Rusch},
  \bibinfo{person}{Marc Pickett}, \bibinfo{person}{Kathleen Meier-Hellstern},
  \bibinfo{person}{Meredith~Ringel Morris}, \bibinfo{person}{Tulsee Doshi},
  \bibinfo{person}{Renelito~Delos Santos}, \bibinfo{person}{Toju Duke},
  \bibinfo{person}{Johnny Soraker}, \bibinfo{person}{Ben Zevenbergen},
  \bibinfo{person}{Vinodkumar Prabhakaran}, \bibinfo{person}{Mark Diaz},
  \bibinfo{person}{Ben Hutchinson}, \bibinfo{person}{Kristen Olson},
  \bibinfo{person}{Alejandra Molina}, \bibinfo{person}{Erin Hoffman-John},
  \bibinfo{person}{Josh Lee}, \bibinfo{person}{Lora Aroyo},
  \bibinfo{person}{Ravi Rajakumar}, \bibinfo{person}{Alena Butryna},
  \bibinfo{person}{Matthew Lamm}, \bibinfo{person}{Viktoriya Kuzmina},
  \bibinfo{person}{Joe Fenton}, \bibinfo{person}{Aaron Cohen},
  \bibinfo{person}{Rachel Bernstein}, \bibinfo{person}{Ray Kurzweil},
  \bibinfo{person}{Blaise Aguera-Arcas}, \bibinfo{person}{Claire Cui},
  \bibinfo{person}{Marian Croak}, \bibinfo{person}{Ed Chi}, {and}
  \bibinfo{person}{Quoc Le}.} \bibinfo{year}{2022}\natexlab{}.
\newblock \bibinfo{title}{LaMDA: Language Models for Dialog Applications}.
\newblock
\newblock
\showeprint[arxiv]{2201.08239}~[cs.CL]


\bibitem[\protect\citeauthoryear{Triantafillou, Zemel, and
  Urtasun}{Triantafillou et~al\mbox{.}}{2017}]%
        {Triantafillou2017FewShotIRLens}
\bibfield{author}{\bibinfo{person}{Eleni Triantafillou},
  \bibinfo{person}{Richard Zemel}, {and} \bibinfo{person}{Raquel Urtasun}.}
  \bibinfo{year}{2017}\natexlab{}.
\newblock \showarticletitle{Few-Shot Learning through an Information Retrieval
  Lens}. In \bibinfo{booktitle}{\emph{Proceedings of the 31st International
  Conference on Neural Information Processing Systems}} (Long Beach,
  California, USA) \emph{(\bibinfo{series}{NIPS'17})}.
  \bibinfo{publisher}{Curran Associates Inc.}, \bibinfo{address}{Red Hook, NY,
  USA}, \bibinfo{pages}{2252–2262}.
\newblock
\showISBNx{9781510860964}


\bibitem[\protect\citeauthoryear{Vaswani, Shazeer, Parmar, Uszkoreit, Jones,
  Gomez, Kaiser, and Polosukhin}{Vaswani et~al\mbox{.}}{2017}]%
        {vaswani:transformer}
\bibfield{author}{\bibinfo{person}{Ashish Vaswani}, \bibinfo{person}{Noam
  Shazeer}, \bibinfo{person}{Niki Parmar}, \bibinfo{person}{Jakob Uszkoreit},
  \bibinfo{person}{Llion Jones}, \bibinfo{person}{Aidan~N Gomez},
  \bibinfo{person}{\L~ukasz Kaiser}, {and} \bibinfo{person}{Illia Polosukhin}.}
  \bibinfo{year}{2017}\natexlab{}.
\newblock \showarticletitle{Attention is All you Need}.
\newblock In \bibinfo{booktitle}{\emph{Advances in Neural Information
  Processing Systems 30}}, \bibfield{editor}{\bibinfo{person}{I.~Guyon},
  \bibinfo{person}{U.~V. Luxburg}, \bibinfo{person}{S.~Bengio},
  \bibinfo{person}{H.~Wallach}, \bibinfo{person}{R.~Fergus},
  \bibinfo{person}{S.~Vishwanathan}, {and} \bibinfo{person}{R.~Garnett}}
  (Eds.). \bibinfo{publisher}{Curran Associates, Inc.},
  \bibinfo{pages}{5998--6008}.
\newblock


\bibitem[\protect\citeauthoryear{Verga, Sun, Baldini~Soares, and Cohen}{Verga
  et~al\mbox{.}}{2021}]%
        {Verga+al:2021}
\bibfield{author}{\bibinfo{person}{Pat Verga}, \bibinfo{person}{Haitian Sun},
  \bibinfo{person}{Livio Baldini~Soares}, {and} \bibinfo{person}{William
  Cohen}.} \bibinfo{year}{2021}\natexlab{}.
\newblock \showarticletitle{Adaptable and Interpretable Neural {M}emory{O}ver
  Symbolic Knowledge}. In \bibinfo{booktitle}{\emph{Proceedings of the 2021
  Conference of the North American Chapter of the Association for Computational
  Linguistics: Human Language Technologies}}. \bibinfo{publisher}{Association
  for Computational Linguistics}, \bibinfo{address}{Online},
  \bibinfo{pages}{3678--3691}.
\newblock
\urldef\tempurl%
\url{https://doi.org/10.18653/v1/2021.naacl-main.288}
\showDOI{\tempurl}


\bibitem[\protect\citeauthoryear{Weston, Chopra, and Bordes}{Weston
  et~al\mbox{.}}{2015}]%
        {weston:memory-networks}
\bibfield{author}{\bibinfo{person}{Jason Weston}, \bibinfo{person}{Sumit
  Chopra}, {and} \bibinfo{person}{Antoine Bordes}.}
  \bibinfo{year}{2015}\natexlab{}.
\newblock \showarticletitle{Memory Networks}. In
  \bibinfo{booktitle}{\emph{Proceedings of the International Conference on
  Learning Representations (ICLR)}}.
\newblock


\bibitem[\protect\citeauthoryear{Wu, Feichtenhofer, Fan, He, Krahenbuhl, and
  Girshick}{Wu et~al\mbox{.}}{2019a}]%
        {wu2019long}
\bibfield{author}{\bibinfo{person}{Chao-Yuan Wu}, \bibinfo{person}{Christoph
  Feichtenhofer}, \bibinfo{person}{Haoqi Fan}, \bibinfo{person}{Kaiming He},
  \bibinfo{person}{Philipp Krahenbuhl}, {and} \bibinfo{person}{Ross Girshick}.}
  \bibinfo{year}{2019}\natexlab{a}.
\newblock \showarticletitle{Long-term feature banks for detailed video
  understanding}. In \bibinfo{booktitle}{\emph{Proceedings of the IEEE/CVF
  Conference on Computer Vision and Pattern Recognition}}.
  \bibinfo{pages}{284--293}.
\newblock


\bibitem[\protect\citeauthoryear{Wu, Li, Mangalam, Fan, Xiong, Malik, and
  Feichtenhofer}{Wu et~al\mbox{.}}{2022}]%
        {wu2022memvit}
\bibfield{author}{\bibinfo{person}{Chao-Yuan Wu}, \bibinfo{person}{Yanghao Li},
  \bibinfo{person}{Karttikeya Mangalam}, \bibinfo{person}{Haoqi Fan},
  \bibinfo{person}{Bo Xiong}, \bibinfo{person}{Jitendra Malik}, {and}
  \bibinfo{person}{Christoph Feichtenhofer}.} \bibinfo{year}{2022}\natexlab{}.
\newblock \showarticletitle{MeMViT: Memory-Augmented Multiscale Vision
  Transformer for Efficient Long-Term Video Recognition}.
\newblock \bibinfo{journal}{\emph{arXiv preprint arXiv:2201.08383}}
  (\bibinfo{year}{2022}).
\newblock


\bibitem[\protect\citeauthoryear{Wu, Wang, and Wang}{Wu et~al\mbox{.}}{2019b}]%
        {Wu+al:2019}
\bibfield{author}{\bibinfo{person}{Jiawei Wu}, \bibinfo{person}{Xin Wang},
  {and} \bibinfo{person}{William~Yang Wang}.} \bibinfo{year}{2019}\natexlab{b}.
\newblock \showarticletitle{Extract and Edit: An Alternative to
  Back-Translation for Unsupervised Neural Machine Translation}. In
  \bibinfo{booktitle}{\emph{Proceedings of the 2019 Conference of the North
  {A}merican Chapter of the Association for Computational Linguistics: Human
  Language Technologies, Volume 1 (Long and Short Papers)}}.
  \bibinfo{publisher}{Association for Computational Linguistics},
  \bibinfo{address}{Minneapolis, Minnesota}, \bibinfo{pages}{1173--1183}.
\newblock
\urldef\tempurl%
\url{https://doi.org/10.18653/v1/N19-1120}
\showDOI{\tempurl}


\bibitem[\protect\citeauthoryear{Xiong, Xiong, Li, Tang, Liu, Bennett, Ahmed,
  and Overwijk}{Xiong et~al\mbox{.}}{2021}]%
        {Xiong2021ANCE}
\bibfield{author}{\bibinfo{person}{Lee Xiong}, \bibinfo{person}{Chenyan Xiong},
  \bibinfo{person}{Ye Li}, \bibinfo{person}{Kwok-Fung Tang},
  \bibinfo{person}{Jialin Liu}, \bibinfo{person}{Paul~N. Bennett},
  \bibinfo{person}{Junaid Ahmed}, {and} \bibinfo{person}{Arnold Overwijk}.}
  \bibinfo{year}{2021}\natexlab{}.
\newblock \showarticletitle{Approximate Nearest Neighbor Negative Contrastive
  Learning for Dense Text Retrieval}. In
  \bibinfo{booktitle}{\emph{International Conference on Learning
  Representations}} \emph{(\bibinfo{series}{ICLR'21})}.
\newblock


\bibitem[\protect\citeauthoryear{Xu, Ghosh, Huang, Okhonko, Aghajanyan, Metze,
  Zettlemoyer, and Feichtenhofer}{Xu et~al\mbox{.}}{2021a}]%
        {xu-etal-2021-videoclip}
\bibfield{author}{\bibinfo{person}{Hu Xu}, \bibinfo{person}{Gargi Ghosh},
  \bibinfo{person}{Po-Yao Huang}, \bibinfo{person}{Dmytro Okhonko},
  \bibinfo{person}{Armen Aghajanyan}, \bibinfo{person}{Florian Metze},
  \bibinfo{person}{Luke Zettlemoyer}, {and} \bibinfo{person}{Christoph
  Feichtenhofer}.} \bibinfo{year}{2021}\natexlab{a}.
\newblock \showarticletitle{{V}ideo{CLIP}: Contrastive Pre-training for
  Zero-shot Video-Text Understanding}. In \bibinfo{booktitle}{\emph{Proceedings
  of the 2021 Conference on Empirical Methods in Natural Language Processing}}.
  \bibinfo{publisher}{Association for Computational Linguistics},
  \bibinfo{address}{Online and Punta Cana, Dominican Republic},
  \bibinfo{pages}{6787--6800}.
\newblock
\urldef\tempurl%
\url{https://doi.org/10.18653/v1/2021.emnlp-main.544}
\showDOI{\tempurl}


\bibitem[\protect\citeauthoryear{Xu and Croft}{Xu and Croft}{1996}]%
        {Xu:1996}
\bibfield{author}{\bibinfo{person}{Jinxi Xu} {and} \bibinfo{person}{W.~Bruce
  Croft}.} \bibinfo{year}{1996}\natexlab{}.
\newblock \showarticletitle{Query Expansion Using Local and Global Document
  Analysis}. In \bibinfo{booktitle}{\emph{Proceedings of the 19th Annual
  International ACM SIGIR Conference on Research and Development in Information
  Retrieval}} (Zurich, Switzerland) \emph{(\bibinfo{series}{SIGIR '96})}.
  \bibinfo{publisher}{Association for Computing Machinery},
  \bibinfo{address}{New York, NY, USA}, \bibinfo{pages}{4–11}.
\newblock
\showISBNx{0897917928}
\urldef\tempurl%
\url{https://doi.org/10.1145/243199.243202}
\showDOI{\tempurl}


\bibitem[\protect\citeauthoryear{Xu, Guo, Wang, Li, Zhou, and Loy}{Xu
  et~al\mbox{.}}{2021b}]%
        {xu2021texture}
\bibfield{author}{\bibinfo{person}{Rui Xu}, \bibinfo{person}{Minghao Guo},
  \bibinfo{person}{Jiaqi Wang}, \bibinfo{person}{Xiaoxiao Li},
  \bibinfo{person}{Bolei Zhou}, {and} \bibinfo{person}{Chen~Change Loy}.}
  \bibinfo{year}{2021}\natexlab{b}.
\newblock \showarticletitle{Texture memory-augmented deep patch-based image
  inpainting}.
\newblock \bibinfo{journal}{\emph{IEEE Transactions on Image Processing}}
  \bibinfo{volume}{30} (\bibinfo{year}{2021}), \bibinfo{pages}{9112--9124}.
\newblock


\bibitem[\protect\citeauthoryear{Yang, Hu, Qiu, Qu, Gao, Croft, Liu, Shen, and
  Liu}{Yang et~al\mbox{.}}{2019}]%
        {Yang2019RetrGenConv}
\bibfield{author}{\bibinfo{person}{Liu Yang}, \bibinfo{person}{Junjie Hu},
  \bibinfo{person}{Minghui Qiu}, \bibinfo{person}{Chen Qu},
  \bibinfo{person}{Jianfeng Gao}, \bibinfo{person}{W.~Bruce Croft},
  \bibinfo{person}{Xiaodong Liu}, \bibinfo{person}{Yelong Shen}, {and}
  \bibinfo{person}{Jingjing Liu}.} \bibinfo{year}{2019}\natexlab{}.
\newblock \showarticletitle{A Hybrid Retrieval-Generation Neural Conversation
  Model}. In \bibinfo{booktitle}{\emph{Proceedings of the 28th ACM
  International Conference on Information and Knowledge Management}} (Beijing,
  China) \emph{(\bibinfo{series}{CIKM '19})}. \bibinfo{publisher}{Association
  for Computing Machinery}, \bibinfo{address}{New York, NY, USA},
  \bibinfo{pages}{1341–1350}.
\newblock
\showISBNx{9781450369763}
\urldef\tempurl%
\url{https://doi.org/10.1145/3357384.3357881}
\showDOI{\tempurl}


\bibitem[\protect\citeauthoryear{Zamani and Croft}{Zamani and Croft}{2017}]%
        {Zamani2017RelWE}
\bibfield{author}{\bibinfo{person}{Hamed Zamani} {and}
  \bibinfo{person}{W.~Bruce Croft}.} \bibinfo{year}{2017}\natexlab{}.
\newblock \showarticletitle{Relevance-Based Word Embedding}. In
  \bibinfo{booktitle}{\emph{Proceedings of the 40th International ACM SIGIR
  Conference on Research and Development in Information Retrieval}} (Shinjuku,
  Tokyo, Japan) \emph{(\bibinfo{series}{SIGIR '17})}.
  \bibinfo{publisher}{Association for Computing Machinery},
  \bibinfo{address}{New York, NY, USA}, \bibinfo{pages}{505–514}.
\newblock
\showISBNx{9781450350228}
\urldef\tempurl%
\url{https://doi.org/10.1145/3077136.3080831}
\showDOI{\tempurl}


\bibitem[\protect\citeauthoryear{Zamani and Croft}{Zamani and Croft}{2018}]%
        {Zamani:2018:TheoryWS}
\bibfield{author}{\bibinfo{person}{Hamed Zamani} {and}
  \bibinfo{person}{W.~Bruce Croft}.} \bibinfo{year}{2018}\natexlab{}.
\newblock \showarticletitle{On the Theory of Weak Supervision for Information
  Retrieval}. In \bibinfo{booktitle}{\emph{Proceedings of the 2018 ACM SIGIR
  International Conference on Theory of Information Retrieval}} (Tianjin,
  China) \emph{(\bibinfo{series}{ICTIR '18})}. \bibinfo{publisher}{Association
  for Computing Machinery}, \bibinfo{address}{New York, NY, USA},
  \bibinfo{pages}{147–154}.
\newblock
\showISBNx{9781450356565}
\urldef\tempurl%
\url{https://doi.org/10.1145/3234944.3234968}
\showDOI{\tempurl}


\bibitem[\protect\citeauthoryear{Zhai and Lafferty}{Zhai and Lafferty}{2001}]%
        {Zhai2001Mix}
\bibfield{author}{\bibinfo{person}{Chengxiang Zhai} {and} \bibinfo{person}{John
  Lafferty}.} \bibinfo{year}{2001}\natexlab{}.
\newblock \showarticletitle{Model-Based Feedback in the Language Modeling
  Approach to Information Retrieval}. In \bibinfo{booktitle}{\emph{Proceedings
  of the Tenth International Conference on Information and Knowledge
  Management}} (Atlanta, Georgia, USA) \emph{(\bibinfo{series}{CIKM '01})}.
  \bibinfo{publisher}{Association for Computing Machinery},
  \bibinfo{address}{New York, NY, USA}, \bibinfo{pages}{403–410}.
\newblock
\showISBNx{1581134363}
\urldef\tempurl%
\url{https://doi.org/10.1145/502585.502654}
\showDOI{\tempurl}


\bibitem[\protect\citeauthoryear{Zhang, Bengio, Hardt, Recht, and
  Vinyals}{Zhang et~al\mbox{.}}{2017}]%
        {zhang:noise-memorization}
\bibfield{author}{\bibinfo{person}{Chiyuan Zhang}, \bibinfo{person}{Samy
  Bengio}, \bibinfo{person}{Moritz Hardt}, \bibinfo{person}{Benjamin Recht},
  {and} \bibinfo{person}{Oriol Vinyals}.} \bibinfo{year}{2017}\natexlab{}.
\newblock \showarticletitle{Understanding deep learning requires rethinking
  generalization}. In \bibinfo{booktitle}{\emph{5th International Conference on
  Learning Representations, {ICLR} 2017, Toulon, France, April 24-26, 2017,
  Conference Track Proceedings}}. \bibinfo{publisher}{OpenReview.net}.
\newblock
\urldef\tempurl%
\url{https://openreview.net/forum?id=Sy8gdB9xx}
\showURL{%
\tempurl}


\bibitem[\protect\citeauthoryear{Zhang, Sun, Gao, Fang, Brockett, Galley, Gao,
  and Dolan}{Zhang et~al\mbox{.}}{2022}]%
        {zhangAAAI22}
\bibfield{author}{\bibinfo{person}{Yizhe Zhang}, \bibinfo{person}{Siqi Sun},
  \bibinfo{person}{Xiang Gao}, \bibinfo{person}{Yuwei Fang},
  \bibinfo{person}{Chris Brockett}, \bibinfo{person}{Michel Galley},
  \bibinfo{person}{Jianfeng Gao}, {and} \bibinfo{person}{Bill Dolan}.}
  \bibinfo{year}{2022}\natexlab{}.
\newblock \showarticletitle{Joint Retrieval and Generation Training for
  Grounded Text Generation}. In \bibinfo{booktitle}{\emph{AAAI}}.
\newblock


\bibitem[\protect\citeauthoryear{Zhu, Lei, Wang, Zheng, Poria, and Chua}{Zhu
  et~al\mbox{.}}{2021}]%
        {zhu2021retrieving}
\bibfield{author}{\bibinfo{person}{Fengbin Zhu}, \bibinfo{person}{Wenqiang
  Lei}, \bibinfo{person}{Chao Wang}, \bibinfo{person}{Jianming Zheng},
  \bibinfo{person}{Soujanya Poria}, {and} \bibinfo{person}{Tat-Seng Chua}.}
  \bibinfo{year}{2021}\natexlab{}.
\newblock \showarticletitle{Retrieving and reading: A comprehensive survey on
  open-domain question answering}.
\newblock \bibinfo{journal}{\emph{arXiv preprint arXiv:2101.00774}}
  (\bibinfo{year}{2021}).
\newblock


\end{thebibliography}
\end{document}